\documentclass{article}
\usepackage{iclr2027_conference,times}

\usepackage{amsmath,amsfonts,bm}

\def\eqref#1{equation~\ref{#1}}
\def\Eqref#1{Equation~\ref{#1}}

\def\1{\bm{1}}

\DeclareMathAlphabet{\mathsfit}{\encodingdefault}{\sfdefault}{m}{sl}
\SetMathAlphabet{\mathsfit}{bold}{\encodingdefault}{\sfdefault}{bx}{n}

\newcommand{\E}{\mathbb{E}}

\newcommand{\R}{\mathbb{R}}

\newcommand{\KL}{D_{\mathrm{KL}}}

\DeclareMathOperator*{\argmin}{arg\,min}

\usepackage{amsmath,amssymb,amsthm,mathtools}
\usepackage{booktabs,multirow}
\usepackage{algorithm}
\usepackage[noend]{algpseudocode}
\usepackage{graphicx}
\usepackage{placeins}
\usepackage{tikz}
\usetikzlibrary{arrows.meta,calc}
\usepackage{xcolor}
\usepackage{microtype}
\usepackage{enumitem}
\usepackage{hyperref}
\hypersetup{hidelinks,pdftitle={Elysium: Simulation-Free Unbalanced Schrodinger Bridges}}
\usepackage{url}

\newtheorem{theorem}{Theorem}
\newtheorem{proposition}[theorem]{Proposition}
\newtheorem*{proposition*}{Proposition}
\newtheorem{corollary}[theorem]{Corollary}

\theoremstyle{definition}
\newtheorem{definition}[theorem]{Definition}
\theoremstyle{plain}
\newcommand{\method}{\textsc{Elysium}}
\newcommand{\state}{\mathcal{X}}
\newcommand{\coffin}{\dagger}
\newcommand{\Normal}{\mathcal{N}}
\newcommand{\dd}{\mathrm{d}}
\newcommand{\rev}[1]{#1}
\newenvironment{revised}{}{}
\makeatletter
\let\rawtableinput\@@input
\makeatother

\title{\raggedright Elysium: Simulation-Free\\
Unbalanced Schr\"odinger Bridges}

\author{Zihan Wang\textsuperscript{1}\thanks{Equal contribution.}\,,
  Jianzhe Li\textsuperscript{2}\footnotemark[1]\,,
  Zhenyi Zhang\textsuperscript{3}\thanks{Corresponding authors: \texttt{zhenyizhang@stu.pku.edu.cn}, \texttt{pjzhou@pku.edu.cn}.}\ \ \&
  Peijie Zhou\textsuperscript{1,2,4,5}\footnotemark[2] \\
  \\
  \textsuperscript{1}Center for Quantitative Biology, Peking University \\
  \textsuperscript{2}Center for Machine Learning Research, Peking University \\
  \textsuperscript{3}LMAM and School of Mathematical Sciences, Peking University \\
  \textsuperscript{4}National Engineering Laboratory for Big Data Analysis and Applications, Beijing \\
  \textsuperscript{5}AI for Science Institute, Beijing
}

\iclrfinalcopy

\begin{document}
\maketitle
\raggedbottom
\setlength{\textfloatsep}{9pt plus 2pt minus 2pt}
\setlength{\floatsep}{7pt plus 2pt minus 2pt}
\setlength{\intextsep}{9pt plus 2pt minus 2pt}
\setlength{\abovecaptionskip}{5pt}
\makeatletter
\setlength{\@fptop}{0pt}
\makeatother

\begin{abstract}
Reconstructing cellular dynamics from snapshots is challenging because the
destructive nature of single-cell measurements makes tracking the same
cells over time impossible.  Cellular dynamics are inherently stochastic
and involve changes in population size through proliferation and
apoptosis.  Growth-based methods represent population change as a rate
on the density, which does not assign exits and entries to individual cells.
Unbalanced Schr\"odinger bridges model individual events, but neural solvers
for killed diffusion bridges simulate trajectories during training.  We
introduce \method{}, a
simulation-free solver that jointly
learns state transition and mass change.  Using a killed Brownian reference
and its time reversal, \method{} constructs direct regression objectives
from an optimal endpoint coupling and conditional bridge distributions,
without simulating trajectories during training.  We show that the
minimizers of the expected training objectives recover the optimal
unbalanced Schr\"odinger bridge.  Empirically, \method{} reconstructs
state transitions, population size, and event distributions on simulated benchmarks with known
killed-diffusion ground truth, outperforming existing deterministic and bridge-based solvers.  On
single-cell time courses, it interpolates cell state distributions at
unobserved time points, tracks population expansion and contraction, and captures biphasic drug-response dynamics.
\end{abstract}

\section{Introduction}
\label{sec:introduction}

Population dynamics are often observed through unpaired snapshots that
sample individual states without linking the same individuals across
times.  In single-cell biology, measurements destroy the cells, so the same
cells cannot be tracked \citep{schiebinger2019optimal,weinreb2018fundamental},
and proliferation and apoptosis change the population size.  Our goal is to
learn, from snapshots alone, a continuous-time stochastic process that
jointly captures state transitions and population changes.

Optimal transport couples consecutive snapshots, and flow matching turns
such couplings into continuous dynamics by regressing a velocity field on
sampled intermediate states
\citep{benamou2000computational,peyre2019computational,lipman2022flow,tong2023minibatch,tong2020trajectorynet}.
Unbalanced optimal transport extends this framework to changing population
sizes through a reaction term, as in the Wasserstein--Fisher--Rao (WFR)
geometry \citep{chizat2018scaling,liero2018optimal}, and WFR-based solvers
learn a velocity field together with a growth rate
\citep{wang2025vgfm,peng2026wfrfm,sha2024tigon}.  These models describe the
population through its density, and a reaction field records only the net
gain or loss at each location.  The same net change can arise from different
combinations of individual entry and exit events.  The trajectories are
also deterministic, so individuals starting from the same state cannot diverge.
\rev{Representing individuals} requires a stochastic process over individual histories in
which exits and entries are events\rev{, although which combination of events
occurred cannot be read from snapshots alone
(Appendix~\ref{app:identifiability})}.

Schr\"odinger bridges (SBs) provide such a process.  A reference
stochastic process encodes a prior over individual behavior, and the
bridge is the process closest to the reference in relative entropy among
those matching the observed snapshots
\citep{schrodinger1932theorie,leonard2014survey}.  Because the prior acts
on complete histories, events in the reference become events of individuals
in the bridge.  Under a killed-diffusion reference, where each individual
diffuses and exits at some rate, the bridge assigns trajectories,
exit times, and exit locations, and reversing time turns exits into
entries on growing intervals
\citep{chen2022most,chen2025survival,pariset2023unbalanced}.

The challenge is to learn this bridge without simulating trajectories
during training.  UDSB uses neural iterative proportional fitting with
alternating forward and backward simulations
\citep{debortoli2021dsb,pariset2023unbalanced}.  For balanced bridges, bridge
matching removes simulation by regressing on conditional reference paths
\citep{tong2024simulationfree}.  USB \citep{ying2026beyond} brings
simulation-free training to a branching SB, but it approximates the coupling
with a WFR problem and regresses a signed growth rate, from which division or
death is simulated after training.  No existing solver obtains simulation-free targets for
motion, exits, and entries directly from an unbalanced SB.

We introduce \method{}, a neural solver with simulation-free training for the killed-diffusion unbalanced SB.  Building on the cemetery-state formulation
\citep{chen2022most,chen2025survival,pariset2023unbalanced}, \method{} combines
an exact endpoint projection with analytic conditional path laws to
construct direct training targets (Figure~\ref{fig:method-overview}).  Given the endpoint coupling, the optimal bridge admits a closed-form decomposition into survival, exit, and entry components. Survivors follow Brownian bridges, exiting individuals follow Brownian motion with a conditioned exit clock, and time reversal yields the corresponding entry component. These analytic component laws provide simulation-free regression targets for the drift and exit rate, while a generative flow learns the coupling-weighted endpoint distribution for entry locations. At the population optimum, the learned generator recovers the path law of the optimal bridge.  Across simulated benchmarks with known killed-diffusion ground truth and real single-cell time courses, \method{} reconstructs unobserved intermediate state distributions, population size trajectories, and drug-induced expansion--contraction dynamics without simulating trajectories during training.

\section{Related work}
\label{sec:related}

\paragraph{Optimal transport and flow matching.}
Dynamic OT represents distribution evolution by a velocity field
\citep{benamou2000computational,peyre2019computational}, and flow matching
learns this field through conditional regression
\citep{lipman2022flow,tong2023minibatch}.  TrajectoryNet and PRESCIENT
apply neural dynamical models to single-cell snapshot data
\citep{tong2020trajectorynet,yeo2021prescient}.  Unbalanced OT augments
transport with reaction, as in the WFR geometry
\citep{chizat2018scaling,liero2018optimal}.  Continuous population solvers including Action Matching, WLF, VGFM,
TIGON, DeepRUOT, and WFR-FM learn velocity and growth dynamics under
such objectives
\citep{neklyudov2022action,neklyudov2023wlf,wang2025vgfm,sha2024tigon,zhang2025deepruot,peng2026wfrfm},
while SF-EUOT relaxes the terminal marginal of entropic OT to admit mass
change \citep{choi2024sfeuot}.  All of these represent mass change through
a growth field or a relaxed constraint on the density.  \method{} models
motion, exits, and entries as mechanisms of one stochastic path law, so
every unit of created or removed mass has a state, a time, and a location.

\paragraph{Schr\"odinger bridges with changing mass.}
SBs project a reference path law onto prescribed marginals
\citep{schrodinger1932theorie,leonard2014survey}.  Balanced neural solvers
typically alternate forward and backward simulations
\citep{debortoli2021dsb,shi2024dsbm}, and bridge matching removes
simulation by regressing on conditional reference paths
\citep{tong2024simulationfree}.  With changing mass, killed SBs were
formulated and solved with Sinkhorn-type iterations by
\citet{chen2022most,chen2025survival}, and UDSB combines exit with
reverse-time entry in a neural iterative proportional fitting scheme that
again alternates simulations \citep{pariset2023unbalanced}.  BranchSBM
and USB model branching and reproduction, and CytoBridge models cellular
interactions and proliferation
\citep{baradat2021branching,tang2025branchsbm,zhang2025cytobridge,ying2026beyond}.
Among them, USB avoids simulation by approximating its branching coupling
with a WFR objective.  \method{} derives both the endpoint projection and
the regression targets from one killed Brownian reference
(Appendix~\ref{app:scope}).

\paragraph{Single-cell trajectory inference.}
Destructive assays require trajectories to be reconstructed without cell
correspondence, a problem with fundamental identifiability limits
\citep{schiebinger2019optimal,weinreb2018fundamental,lavenant2024trajectory}.
RNA velocity and lineage tracing add kinetic or ancestral constraints
\citep{lamanno2018rna,bergen2020generalizing,weinreb2020lineage,wang2026phylofm}, and
population balance models constrain growth with additional measurements
\citep{fischer2019inferring}, but many time-course datasets provide only
snapshots and abundance changes \citep{sha2024tigon,bunne2021learning}.
\method{} targets this setting and learns continuous state motion together
with stochastic exits and entries.

\section{From unbalanced snapshots to a Schr\"odinger bridge}
\label{sec:fixed-pool-bridge}
\label{sec:theory}

Given two snapshots of unequal total mass, we seek a stochastic process
in which individuals move, exit, or enter without cell-level correspondence.  We formulate the
unbalanced SB under a killed Brownian reference, lift the endpoints to a balanced
coupling via a cemetery state, and obtain closed-form laws for the conditioned paths
(Figure~\ref{fig:method-overview}, with notation summarized in \rev{Appendix~\ref{app:notation}}).

\begin{figure}[t]
  \centering
  \includegraphics[width=\textwidth]{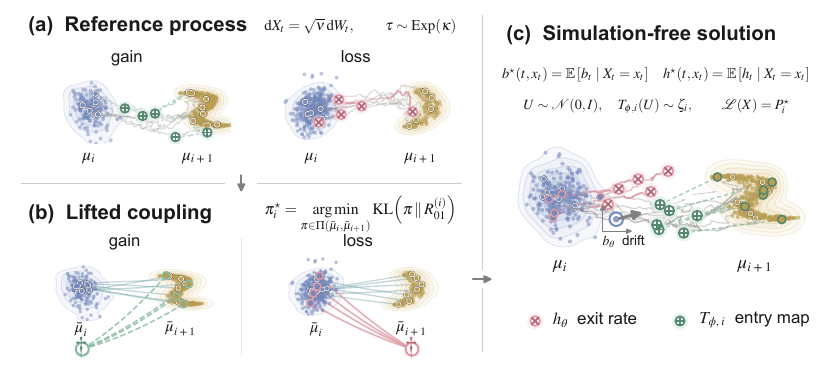}
  \caption{Overview of \method{}.  (a) The killed Brownian reference supplies
  stochastic motion and an exponential exit clock, while reversing the construction
  represents entry on a growing interval.  Shaded contours show snapshot
  densities and particle counts encode mass.  (b) Adding the cemetery balances
  the endpoint measures, and KL projection of the reference endpoint law yields
  a coupling over survival and exit components, with lines denoting endpoint
  pairs.  (c) Closed-form component samples provide
  simulation-free targets for the population drift and exit rate, and the
  entry map transports Gaussian noise to the coupling-weighted endpoint
  distribution \(\zeta_i\) of interval \(i\)
  (Appendix~\ref{app:proof-entry-flow}).  The local arrow shows the learned
  SDE drift at one state.}
  \label{fig:method-overview}
\end{figure}

\subsection{The killed reference and the unbalanced Schr\"odinger bridge}
\label{sec:abstract-usb}
\label{sec:reference}

Rescale one observation interval to \([0,1]\) and let \(\mu_0,\mu_1\) be
the observed finite measures with masses \(m_0\ge m_1\), so that the
observed loss is \(g=m_0-m_1\).  The reference process is Brownian motion
with an independent exponential exit clock\rev{, where each particle carries a random
exit time \(\tau\) satisfying \(\Pr(\tau>t)=e^{-\kappa t}\) independently of
the Brownian motion \(W\)},
\begin{equation}
  \dd X_t=\sqrt{\nu}\,\dd W_t\quad(t<\tau),
  \qquad \tau\sim\mathrm{Exp}(\kappa),
  \qquad X_t=\coffin\quad(t\geq\tau),
  \label{eq:reference-normalized}
\end{equation}
with \(\nu,\kappa>0\)\rev{, and \(R\) denotes the path measure on
\([0,1]\) with initial measure \(\mu_0\) and total mass \(m_0\)}.  \rev{A particle is active while \(t<\tau\).}  An active particle exits
during the next short interval \(\dd t\) with probability \(\kappa\,\dd t\),
and after exiting it sits in an absorbing \emph{cemetery state} \(\coffin\)
and no longer belongs to the active population.  \rev{In a cell population,
an exit is any event that removes a cell from the measured population, such as death, differentiation out of the profiled compartment, or loss in sampling, and an entry is any event that adds one, such as division, influx, or a higher capture rate
(Appendix~\ref{app:biology}).}  For a growing interval
with \(m_1>m_0\), we swap the snapshots, solve the mass-decreasing
problem in backward time, and reverse the resulting bridge, so it
suffices to treat \(m_0\ge m_1\).  For a path measure \(P\), let
\(P_t^{\mathrm{act}}\) denote the finite measure of particles still active
at time \(t\).

\begin{definition}[Unbalanced Schr\"odinger bridge]
\label{def:usb}
Given \(\mu_0,\mu_1\) with \(m_0\geq m_1\) and the killed reference path
measure \(R\) \rev{of Equation~\ref{eq:reference-normalized}}, the unbalanced Schr\"odinger bridge minimizes relative
entropy among path measures matching both active endpoints:
\begin{equation}
  P^*=\argmin_{P:\;P_0^{\mathrm{act}}=\mu_0,\;
                       P_1^{\mathrm{act}}=\mu_1}
  \KL(P\Vert R).
  \label{eq:abstract-usb}
\end{equation}
Both \(P\) and \(R\) have mass \(m_0\), and \(\KL\) is relative entropy of
finite measures (Appendix~\ref{app:marked-reference}).
\end{definition}

\rev{Intuitively, \(P^*\) characterizes the most probable evolution of the population
bridging the two snapshots \citep{chen2022most}.}
The problem has an equivalent dynamic form.  Change the drift to
\(\sqrt{\nu}\,u_t\) and the exit rate to \(h_t\ge0\)\rev{, where
\(u_t=u(t,X_t)\) and \(h_t=h(t,X_t)\) are evaluated at the current state of
an active particle}, and write \(P^{u,h}\)
for the resulting path measure \rev{and \(1\wedge\tau\) for \(\min\{1,\tau\}\)}.  Girsanov's theorem gives the following
statement \citep{chen2022most}.

\begin{proposition*}[Equivalent dynamic formulation]
Under the conditions in Appendix~\ref{app:abstract-usb-proof},
Equation~\ref{eq:abstract-usb} has the same value and the same optimal
process as
\begin{equation}
 \inf_{u,h}\;
 \E_{P^{u,h}}\!\int_0^{1\wedge\tau}
 \left[
   \frac12\lVert u_t\rVert^2
   +h_t\log\frac{h_t}{\kappa}
   -h_t+\kappa
 \right]\dd t
 \quad\text{subject to}\quad
 P_0^{\mathrm{act}}=\mu_0,\;
 P_1^{\mathrm{act}}=\mu_1.
\label{eq:dynamic-kl-main}
\end{equation}
\end{proposition*}

The quadratic term penalizes changes to the reference drift, and the
second term is the relative-entropy rate of the exit mechanism, which
vanishes when \(h_t=\kappa\).  Both costs accumulate only while a particle
is active, so the dynamic problem controls motion and exit timing while
requiring the final active distribution to match the target snapshot.
Solving this control problem directly would require simulating controlled
trajectories during training, which we avoid through the endpoint reduction of Section~\ref{sec:cemetery-lift}.

\subsection{From path optimization to endpoint coupling}
\label{sec:cemetery-lift}

In the classical Schr\"odinger bridge, conditioning on the initial and
final states separates path-space relative entropy into an endpoint term
and a conditional-path term \citep{leonard2014survey}.  Let \(\pi=P_{01}\)
and \(R_{01}\) be the endpoint couplings of \(P\) and \(R\), and let
\(P^{xy},R^{xy}\) be their path laws conditioned on the endpoint pair
\((x,y)\).  The KL chain rule gives
\begin{equation}
 \KL(P\Vert R)=\KL(\pi\Vert R_{01})
 +\int\KL(P^{xy}\Vert R^{xy})\,\pi(\dd x,\dd y).
 \label{eq:path-kl-chain-rule-main}
\end{equation}
For a fixed coupling \(\pi\), choosing \(P^{xy}=R^{xy}\) makes the second
term zero, so only the endpoint coupling needs to be optimized and the
conditional paths remain those of the reference.

The chain rule requires every path to have an endpoint at time one, but
an exiting particle has none in the state space.  The cemetery state
provides one.  Recording all
post-exit states as \(\coffin\), with unit point mass \(\delta_{\coffin}\),
the complete endpoint measures become
\begin{equation}
  \mu_0,\qquad \mu_1+g\delta_{\coffin},
  \label{eq:abstract-lift}
\end{equation}
and both have mass \(m_0\).  This representation changes only how
histories are recorded and preserves the constraints and the
relative-entropy cost, so matching the active endpoints \((\mu_0,\mu_1)\)
and matching the complete endpoints
\((\mu_0,\mu_1+g\delta_{\coffin})\) have the same optimal motion and exit
process.  Below, \(P,R\) denote the extended path measures.

\begin{proposition}[Endpoint reduction]
\label{prop:abstract-usb}
Let \(\Pi(\mu_0,\mu_1+g\delta_{\coffin})\) be the set of couplings with the
marginals in Equation~\ref{eq:abstract-lift}, and suppose it contains a
coupling with finite KL relative to \(R_{01}\).  Under the conditions in
Appendix~\ref{app:abstract-usb-proof}, the optimal path measure of
Definition~\ref{def:usb} is determined by
\begin{equation}
 \pi^*=\argmin_{\pi\in\Pi(\mu_0,\,\mu_1+g\delta_{\coffin})}
 \KL(\pi\Vert R_{01})
 \label{eq:abstract-endpoint-projection}
\end{equation}
and recovered by mixing conditional reference paths,
\begin{equation}
 P^*=\int R^{xy}\,\pi^*(\dd x,\dd y).
 \label{eq:abstract-component-mixture-main}
\end{equation}
\end{proposition}

The total exit mass is fixed at \(g\), while the coupling determines which
source mass exits.  For the reference in
Equation~\ref{eq:reference-normalized}, independence of the Brownian
motion and the exit clock gives the endpoint law
\begin{equation}
 R_{01}(\dd x,\dd y)=\mu_0(\dd x)
 \left[e^{-\kappa}\varphi_\nu(y-x)\,\dd y
 +(1-e^{-\kappa})\delta_{\coffin}(\dd y)\right],
 \label{eq:killed-kernel}
\end{equation}
where \(\varphi_\nu\) is the density of \(\Normal(0,\nu I)\).  A particle
survives to the endpoint with probability \(e^{-\kappa}\) and undergoes a
Gaussian displacement, or exits with probability \(1-e^{-\kappa}\) and
ends at \(\coffin\).  Up to constants, which do not affect the coupling because the marginals
fix the mass of each branch, the survival branch has quadratic cost and the
exit branch has position-independent cost.
Equation~\ref{eq:abstract-endpoint-projection} is therefore an entropic
optimal transport problem on the extended endpoints and is solved with
Sinkhorn iterations \citep{cuturi2013sinkhorn,benamou2015bregman}
(Appendices~\ref{app:marked-reference}--\ref{app:killed-kernel-derivation}).
\rev{On samples, the kernel is the Gaussian kernel of entropic OT with one
extra cemetery column.  Without exit, the problem is the static
Schr\"odinger problem, which is entropic OT with cost \(\lVert x-y\rVert^2/(2\nu)\)
\citep{leonard2014survey}, and Sinkhorn is iterative proportional fitting
applied to the endpoint law.  The computed coupling is thus the exact static solution
for the sampled snapshots, so the bridges it defines are learned in a single
regression (Section~\ref{sec:consistency}).  When every interval loses mass, the killed reference splits the
multi-marginal problem exactly into one such projection per adjacent pair, and growing intervals are joined as per-interval bridges
(Appendix~\ref{app:static-structure}).}

\subsection{Closed-form components of the optimal bridge}
\label{sec:components}

Once \(\pi^*\) is known, every conditional reference path in
Equation~\ref{eq:abstract-component-mixture-main} is explicit.  We record
its state law together with the drift and exit rate that generate it,
since these are the quantities the network will regress.

\begin{theorem}[Closed-form bridge decomposition]
\label{thm:reduced-lift}
Let \(\pi^*\) solve Equation~\ref{eq:abstract-endpoint-projection} and let
\(t\in(0,1)\).  All components keep the reference diffusion coefficient
\(\sqrt{\nu}I\).
For a surviving pair \(x\to y\), \(R^{xy}\) is the Brownian bridge from
\(x\) to \(y\), with state law
\(\Normal\big((1-t)x+ty,\nu t(1-t)I\big)\), drift
\(b_t^{xy}(z)=(y-z)/(1-t)\) at active state \(z\), and exit rate
\(h_t^{xy}(z)=0\).
For an exit pair \(x\to\coffin\), \(R^{x\coffin}\) is Brownian motion from
\(x\) whose independent exit clock is conditioned on \(\tau\le1\), and the
particle stays at the cemetery after exit.  While active, the state law is
\(\Normal(x,\nu tI)\), the drift is \(b_t^{x\coffin}(z)=0\), and the exit
rate is
\[
 h_t^{x\coffin}(z)
 =\lim_{\varepsilon\downarrow0}
 \frac{\Pr(t<\tau\leq t+\varepsilon\mid t<\tau\leq1)}{\varepsilon}
 =\frac{\kappa}{1-e^{-\kappa(1-t)}}.
\]
Mixing these conditional paths with the masses of \(\pi^*\) yields the
optimal bridge \(P^*\).  Reversing the mass-decreasing reference and its
optimal bridge gives the bridge of a growing interval, in which an entry
component \(\coffin\to y\) has state law \(\Normal(y,\nu(1-t)I)\) once it
has entered, drift \(b_t^{\coffin y}(z)=(y-z)/(1-t)\), and exit rate zero.
\end{theorem}

The survival drift is the Doob transform of the Brownian transition
density conditioned on the endpoint, whereas conditioning the exit clock on
\(\tau\le1\) leaves the pre-exit motion unchanged and only rescales the
rate.  Proofs are in
Appendices~\ref{app:proof-reduced-lift}--\ref{app:conditional-law-derivations}.

\rev{To illustrate Theorem~\ref{thm:reduced-lift}, consider unit initial mass at \(x\) at \(t=0\) and mass \(0.6\) at \(y\) at
\(t=1\), which fixes the lifted coupling at \(\pi^*(x,y)=0.6\) and
\(\pi^*(x,\coffin)=0.4\).  The surviving mass
\(0.6\) follows the Brownian bridge from \(x\) to \(y\) with drift
\((y-z)/(1-t)\) and no exits, while the remaining \(0.4\) diffuses from \(x\)
without drift and exits at rate \(\kappa/(1-e^{-\kappa(1-t)})\) so that all of it is removed by \(t=1\).  A state at time \(t\) is thus drawn from
one of two Gaussians with explicit component drift and exit rate, providing the regression targets of
Section~\ref{sec:population-recovery}.  On a growing interval from mass
\(0.6\) at \(x\) to mass \(1\) at \(y\), the \(0.6\) follows the same
bridge and the remaining \(0.4\) enters at random times with law
\(\Normal(y,\nu(1-t)I)\) before moving toward \(y\).}

\section{Simulation-free learning of the bridge generator}
\label{sec:learning}

Theorem~\ref{thm:reduced-lift} gives, for every component of the optimal
bridge, a state law together with the drift and exit rate that generate it.  We
now turn these component laws into regression targets for the population
drift and exit rate, model the entry source on growing intervals,
and show that the learned generator recovers \(P^*\).

\subsection{Learning state motion and exit events}
\label{sec:population-recovery}

The component coefficients depend on the endpoint pair, which is unknown
when generating from an initial population.  We therefore learn a drift and
exit rate that depend only on time and the current state, as neural fields \(b_\theta(t,z)\) and
\(h_\theta(t,z)\ge0\), using the component coefficients
\(b_t^c(z),h_t^c(z)\) of Theorem~\ref{thm:reduced-lift} as targets, where
\(c\) indexes the component of an endpoint pair \rev{(\(xy\), \(x\coffin\), or \(\coffin y\))}.

The contribution of a component at time \(t\) is the mass that is still
active.  Survival components remain active throughout the interval.  The
active fraction of an exit component is
\begin{equation}
 s_\kappa(t)
 =\Pr(\tau>t\mid\tau\leq1)
 =\frac{e^{-\kappa t}-e^{-\kappa}}{1-e^{-\kappa}},
 \label{eq:component-mass-factors}
\end{equation}
so for a discrete coupling the survival and exit components have active
masses \(\pi^*_{xy}\) and \(\pi^*_{x\coffin}s_\kappa(t)\).  Growing
intervals use the transposed coupling, and an entry component has entered
with fraction \(s_\kappa(1-t)\).  Normalizing the active masses defines a
component label \(C_t\), the conditional state law of Theorem~\ref{thm:reduced-lift}
defines a state \(Z_t\), and the targets are \(b_t=b_t^{C_t}(Z_t)\) and
\(h_t=h_t^{C_t}(Z_t)\).  A training sample therefore consists of a time,
a component drawn with probability proportional to its active mass, a
state drawn from that component's Gaussian law, and the drift and exit
rate of the component at that state.  All of these are available in closed
form, so no trajectory is simulated during training
(Appendix~\ref{app:conditional-law-derivations}).

We use squared error for the drift and Poisson relative entropy
\(\ell_{\mathrm P}(a,b)=a\log(a/b)-a+b\) \rev{(the same divergence as in Equation~\ref{eq:dynamic-kl-main})} for the exit rate,
\begin{equation}
 \mathcal L_{\mathrm{fields}}(\theta)
 =\E\!\left[
 \lVert b_\theta(t,Z_t)-b_t\rVert^2
 +\ell_{\mathrm P}\!\left(h_t,h_\theta(t,Z_t)\right)
 \right],
 \label{eq:training-loss}
\end{equation}
where the expectation is over training times and the active population
defined above.  The pointwise minimizers of both terms are the conditional
means of \(b_t\) and \(h_t\) given time and current state.  Near a
state where survival and exit components overlap, the regression
therefore averages the survival drift toward the endpoints with the
zero drift of exiting particles, and the exit rate becomes the
conditional rate of the mixture at that state.
Section~\ref{sec:consistency} identifies these conditional means with the
dynamics of the optimal bridge.

\subsection{Constructing and learning the entry source}
\label{sec:entrance}

On a growing interval, Theorem~\ref{thm:reduced-lift} gives the drift of a
particle after it has entered.  It remains to specify when and where
particles enter.  Let \(\pi^*_{\coffin y}\) be the mass of the entry
component \(\coffin\to y\) in the transposed coupling, with total
\(g=\sum_y\pi^*_{\coffin y}>0\).

By Equation~\ref{eq:component-mass-factors}, the fraction of an entry
component that has entered by time \(t\) is \(s_\kappa(1-t)\), so the mass
entering during a short interval \(\Delta t\) is approximately
\(g\,\frac{\dd}{\dd t}s_\kappa(1-t)\,\Delta t\).  Because the exit clock is
independent of the Brownian motion, a particle that enters at time \(t\)
with endpoint \(y\) has state law \(\Normal(y,\nu(1-t)I)\).  Weighting by
the coupling gives the entry source
\begin{equation}
 \eta_t(z)
 =\frac{\dd}{\dd t}s_\kappa(1-t)
  \sum_y\pi^*_{\coffin y}\,
  \varphi_{\nu(1-t)}(z-y),
 \label{eq:entry-source}
\end{equation}
where \(\eta_t(z)\,\dd z\,\dd t\) is the mass entering a volume element
around \(z\) during \(\dd t\).  The entry clock is analytic and the entry
locations follow a Gaussian mixture whose centers are the endpoints
\(y\) weighted by \(\pi^*_{\coffin y}/g\).

Generation must sample entry states without storing the observed
snapshots, so we learn this endpoint distribution with a lightweight
generative flow trained by conditional flow matching.
Draw an endpoint \(Y\) with weights \(\pi^*_{\coffin y}/g\) together with independent
\(U\sim\Normal(0,I)\) and flow time
\(s\sim\mathrm{Unif}[0,1]\).  Along the interpolation \((1-s)U+sY\), a
vector field \(u_\phi\) regresses the velocity \(Y-U\),
\[
 \mathcal L_{\mathrm{entry}}(\phi)
 =\E_{s,U,Y}\left\lVert
 u_\phi\!\left(s,(1-s)U+sY\right)-(Y-U)
 \right\rVert^2.
\]
Integrating \(u_\phi\) from \(s=0\) to \(s=1\) defines a map \(T_\phi\)
from noise to endpoints.  When an entry event occurs at time \(t\), the new
state is
\begin{equation}
 Z_t=T_\phi(U)+\sqrt{\nu(1-t)}\,\epsilon,
 \qquad U,\epsilon\stackrel{\mathrm{iid}}{\sim}\Normal(0,I),
 \label{eq:generated-birth-source}
\end{equation}
so \(T_\phi(U)\) supplies the endpoint mixture center and
the Gaussian perturbation places the particle at its entry position.

\begin{proposition}[Recovery of the entry source]
\label{prop:entrance-amortization}
If \(T_\phi\) matches the coupling-weighted endpoint distribution, then
generating entry events at rate \(g\,\frac{\dd}{\dd t}s_\kappa(1-t)\) with
states drawn from Equation~\ref{eq:generated-birth-source} reproduces the
source in Equation~\ref{eq:entry-source} exactly
(Appendix~\ref{app:proof-entry-flow}).
\end{proposition}

\subsection{The learned generator solves the unbalanced SB}
\label{sec:consistency}

The training samples above are drawn from the optimal bridge \(P^*\).
Conditioning their drift and exit coefficients on the current state
recovers the Markov dynamics of \(P^*\), which connects the regression
objectives back to Definition~\ref{def:usb}.

\begin{theorem}[Exact recovery of the target unbalanced SB]
\label{thm:population-consistency}
Assume the endpoint projection is exact, the regression optima are attained
at almost every interior time, and each entry map matches the endpoint
distribution specified by its coupling.  Under the regularity and
uniqueness-in-law conditions of
Appendix~\ref{app:proof-population-consistency}, the optima are
\begin{equation}
 b^*(t,z)=\E[b_t\mid Z_t=z],
 \qquad h^*(t,z)=\E[h_t\mid Z_t=z].
 \label{eq:population-fields}
\end{equation}
On a decreasing interval, the killed diffusion
\begin{equation}
 \dd X_t=b^*(t,X_t)\dd t+\sqrt{\nu}\,\dd W_t
 \label{eq:sde-inference}
\end{equation}
with exit rate \(h^*\) has path law \(P^*\).  On a growing interval, the
same construction with the exact entry source has the time-reversed
optimal law.  The learned generator therefore solves the target unbalanced
SB and matches all of its active marginals.
\end{theorem}

\rev{The proof is a Markovian projection.  The regression optima average
the component drifts and exit rates over the components active at
\((t,z)\).  Because \(P^*\) reweights the Markov killed reference by
functions of the endpoints alone (a Doob transform), it is itself Markov with exactly these averaged coefficients,
so the killed diffusion they define has law \(P^*\).  In bridge-matching
terms \citep{shi2024dsbm}, Theorem~\ref{thm:reduced-lift} is the
reciprocal projection and Equation~\ref{eq:training-loss} the Markovian
projection.  Started from the exact coupling, one projection suffices (Appendix~\ref{app:static-structure}).  The theorem is the unbalanced counterpart of the guarantee of \citet{tong2023minibatch}, whose flow solves dynamic OT in the zero-noise limit with an exact OT plan and matches the marginals of the balanced SB with an entropic plan.}
Appendix~\ref{app:proof-population-consistency} gives the proof,
Appendix~\ref{app:empirical-marked-consistency} gives conditions under
which empirical bridges converge to the population bridge as the endpoint
measures converge, and Appendix~\ref{app:implementation} describes
finite-sample training.  At the population level the active measure
\(\rho_t\) evolves as
\begin{equation}
 \partial_t\rho_t
 =-\nabla\!\cdot(\rho_t b^*)
 +\frac{\nu}{2}\Delta\rho_t-h^*\rho_t+\eta_t,
 \label{eq:fixed-pool-continuity}
\end{equation}
where the drift and Brownian terms move active states, the exit rate
removes mass, and \(\eta_t\) adds it (\(\eta_t=0\) on a decreasing
interval and equals Equation~\ref{eq:entry-source} on a growing one).
Compared with a WFR model, the single signed growth term is split into an
exit rate and an entry source, each realized by events of individuals.

\section{The \method{} algorithm}
\label{sec:method}
\label{sec:algorithm}

\method{} applies the construction above to every adjacent pair of
snapshots in a time course and shares one neural generator across the
sequence (Algorithms~\ref{alg:elysium-fixed-pool}
and~\ref{alg:elysium-inference}, detailed in
Appendix~\ref{app:implementation}).

\paragraph{Generating stochastic population trajectories.}
For an interval of physical duration \(\Delta_i=t_{i+1}-t_i\), let
\(r=(t-t_i)/\Delta_i\) be local normalized time.  An active state follows
\begin{equation}
 \dd X_t=\frac{b_\theta(r,X_t)}{\Delta_i}\dd t
 +\sqrt{\frac{\nu_i}{\Delta_i}}\,\dd W_t,
 \label{eq:implemented-sde}
\end{equation}
and on a decreasing interval it exits at rate \(h_\theta(r,X_t)/\Delta_i\)
in physical time.  On a growing interval the reversed clock samples entry
times and Equation~\ref{eq:generated-birth-source} samples the new states.
Trajectories are simulated only at generation, because every training update uses
an independent analytic sample and requires no forward or backward
integration.  For evaluation, exits are carried as survival weights, which
define the same law as sampling exit times
(Appendix~\ref{app:numerical-generation}).
Appendix~\ref{app:compute} compares computational cost with WFR-FM and
UDSB-F\rev{.  One fit, including the endpoint coupling and generation,
takes 1.1--2.5 minutes on one GPU, and the coupling at most \(11\%\) of
this time (Appendix~\ref{app:ot-cap})}.

In the algorithms, \(\widehat{\mathcal L}\) denotes a minibatch estimate
of the corresponding loss, the population is represented by weighted
particles \(\mathcal P=\{(x_j,w_j)\}\), \(r\) and \(\delta\) are local
time and step size, \(\operatorname{SDEstep}_i\) evolves particles by
Equation~\ref{eq:implemented-sde}, and \(\mathcal P_{\mathrm{ent}}\)
contains the particles entering during the current step, sampled from
Equations~\ref{eq:entry-source}--\ref{eq:generated-birth-source}
(Appendix~\ref{app:numerical-generation}).

\begin{figure}[t]
\footnotesize
\algrenewcommand\algorithmicrequire{\textbf{Input:}}
\algrenewcommand\algorithmicindent{0.9em}
\noindent
\begin{minipage}[t]{0.485\linewidth}
  \vspace{0pt}
  \begin{algorithm}[H]
    \caption{Simulation-free training}
    \label{alg:elysium-fixed-pool}
    \footnotesize
    \raggedright
    \begin{minipage}[t][13\baselineskip][t]{\linewidth}
      \begin{algorithmic}[1]
        \Require snapshots \(\{(t_i,\mu_i)\}_{i=0}^K\),\newline
          parameters \(\{\nu_i,\kappa_i\}\)
        \State Solve and fix endpoint couplings \(\{\pi_i^*\}\)
        \While{field-training budget remains}
          \State Draw \(i,r\) and \((Z_r,b_r,h_r)\) \rev{as in Sec.~\ref{sec:population-recovery}}
          \State \(\theta\leftarrow\operatorname{AdamW}
            (\theta,\nabla_\theta\widehat{\mathcal L}_{\mathrm{fields}})\)
        \EndWhile
        \For{each growing interval \(i\)}
          \While{entry-training budget remains}
            \State Draw a minibatch as in Section~\ref{sec:entrance}
            \State \(\phi\leftarrow\operatorname{AdamW}
              (\phi,\nabla_\phi\widehat{\mathcal L}_{\mathrm{entry}})\)
          \EndWhile
        \EndFor
      \end{algorithmic}
      \vfill
      \noindent\textbf{Output:} Trained networks
        \(b_\theta,h_\theta,\{u_{\phi,i}\}\).
    \end{minipage}
  \end{algorithm}
\end{minipage}\hfill
\begin{minipage}[t]{0.485\linewidth}
  \vspace{0pt}
  \begin{algorithm}[H]
    \caption{Inference}
    \label{alg:elysium-inference}
    \footnotesize
    \raggedright
    \begin{minipage}[t][13\baselineskip][t]{\linewidth}
      \begin{algorithmic}[1]
        \Require initial particles \(\mathcal P_0\), trained
          networks, interval parameters, steps \(L\)
        \State \(\mathcal P\leftarrow\mathcal P_0,\quad\delta\leftarrow1/L\)
        \For{\(i=0,\ldots,K-1\)}
          \For{\(k=0,\ldots,L-1\)}
            \State \(r\leftarrow k\delta\)
            \State \(\mathcal P\leftarrow
              \operatorname{SDEstep}_i(\mathcal P;r,\delta)\)
            \State \textbf{if loss:}
              \(w_j\leftarrow w_j
              e^{-\int_r^{r+\delta}h_\theta(s,X_s^j)\,\dd s}\)
            \State \textbf{if growth:}
              \(\mathcal P\leftarrow\mathcal P\cup\mathcal P_{\mathrm{ent}}\)
            \State Record \(\widehat\mu\leftarrow\sum_jw_j\delta_{x_j}\)
          \EndFor
        \EndFor
      \end{algorithmic}
      \vfill
      \noindent\textbf{Output:} Stochastic paths and
        populations \(\{\widehat\mu_t\}\).
    \end{minipage}
  \end{algorithm}
\end{minipage}
\vspace{-4pt}
\end{figure}

\FloatBarrier
\section{Experiments}
\label{sec:experiments}

We evaluate whether the solver recovers a specified bridge
against independent ground truth, whether the generator matches and
interpolates single-cell population time courses, and whether entry
and exit capture expansion and contraction under drug treatment.
We compare with four unbalanced baselines covering the main modeling
choices, namely WFR-FM under a WFR growth objective
\citep{peng2026wfrfm}, USB via a WFR-coupled branching bridge
\citep{ying2026beyond}, SF-EUOT with a relaxed entropic OT terminal marginal
\citep{choi2024sfeuot}, and the simulation-based killed-diffusion solver UDSB-F
\citep{pariset2023unbalanced}.
All \(W_1\) scores use the feature space supplied with each benchmark
(lower is better), and \method{} reports mean\(\pm\)SD over five trained
models.  Appendix~\ref{app:evaluation-metrics} defines the metrics and
the rest of Appendix~\ref{app:experiments} gives complete protocols and
results.

\subsection{Exact recovery on killed-diffusion benchmarks}
\label{sec:formal-recovery}

We construct unbalanced Schr\"odinger bridge
problems under killed diffusion references in 1D and 2D, varying endpoint
distributions, diffusion and event rates, and population ratios
\(m_1/m_0\) from \(0.1\) to \(10\).  Each method is trained only on
endpoint populations and tested at intermediate times against independently
computed ground truth.  Each method uses one
configuration selected on a separate development set and fixed across all
cases (Appendix~\ref{app:formal-recovery}).

\begin{figure}[t]
  \centering
  \includegraphics[width=0.192\linewidth,viewport=13.5 74 82.5 151.74,clip]{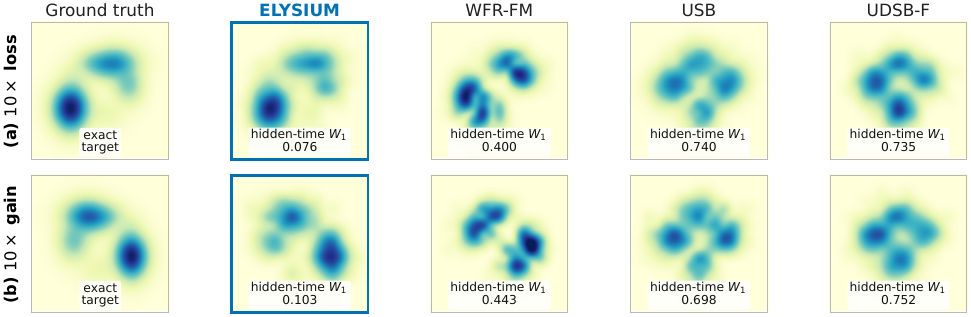}\hfill
  \includegraphics[width=0.192\linewidth,viewport=109.3 74 178.3 151.74,clip]{figures/formal_main_comparison.pdf}\hfill
  \includegraphics[width=0.192\linewidth,viewport=205.1 74 274.1 151.74,clip]{figures/formal_main_comparison.pdf}\hfill
  \includegraphics[width=0.192\linewidth,viewport=300.9 74 369.9 151.74,clip]{figures/formal_main_comparison.pdf}\hfill
  \includegraphics[width=0.192\linewidth,viewport=396.7 74 465.7 151.74,clip]{figures/formal_main_comparison.pdf}
  \caption{Intermediate distribution recovery with \(m_1/m_0=0.1\)
  (population reduced to \(10\%\) of its initial size).  Panels show normalized
  living-population distributions at the midpoint, with labels giving state
  \(W_1\) averaged over the interior test times.
  Appendix~\ref{app:formal-recovery} includes the gain case.}
  \label{fig:formal-main-comparison}
\end{figure}

\begin{table}[t]
  \caption{Mean error across test problems.  State \(W_1\) compares normalized
  living populations, the next two columns measure population size, and joint
  \(W_1\) scores both.  The baseline is the better of USB and WFR-FM.}
  \label{tab:formal-main-summary}
  \centering
  \small
  \renewcommand{\arraystretch}{1.1}
  \setlength{\tabcolsep}{5pt}
  \begin{tabular*}{\linewidth}{@{\extracolsep{\fill}}lcccc@{}}
    \toprule
    Method & State \(W_1\) & Population RME &
    \(\int|\widehat M-M|\) & Joint \(W_1\) \\
    \midrule
    \rawtableinput{tables/formal_main_summary_rows.tex}
    \bottomrule
  \end{tabular*}
\end{table}

\method{} outperforms USB and WFR-FM on all four state-and-mass metrics in
every problem (Table~\ref{tab:formal-main-summary}).  Relative to the better
baseline in each problem, the mean reductions in state \(W_1\) and
integrated mass error are \(75.0\%\) and \(94.2\%\), respectively.
Figure~\ref{fig:formal-main-comparison} illustrates recovery at an unobserved
midpoint with \(m_1/m_0=0.1\), where the baselines distort the modes
that \method{} matches.  On 16 cases with nonzero entry or
exit, each evaluated across five training seeds, \method{} also reduces
event-time \(W_1\) from \(0.1210\) to \(0.0009\) and event-location
\(W_1\) from \(1.1784\) to \(0.0603\) compared with UDSB-F
(Appendix~\ref{app:formal-recovery}).
\rev{For \method{}, the endpoint masses and \(\kappa\) fix the total-mass
curve and the distribution of event times before training
(Equation~\ref{eq:mass-curves}), so state \(W_1\) and event locations
measure what is learned, while total mass and event times check that the
generator reproduces this prior.  Part of the gap to WFR-FM and USB reflects their different objectives, since
even without net exchange (\(m_1/m_0=1\)) their error is already
\(2.0\)--\(4.9\times\) that of \method{}.  The gap grows with imbalance
because the baselines remove or create mass far from the ground-truth
event locations and move mass out of these regions early, which produces
the additional modes in Figure~\ref{fig:formal-main-comparison}
(Appendix~\ref{app:q1-mechanism}).  UDSB-F shares the reference and
endpoint constraints of \method{}, so its gap reflects how the bridge is
learned.  Doubling or quadrupling its training
epochs, or doubling its time steps, changes its mean errors by less than
\(10\%\) (Appendix~\ref{app:udsb-budget}).}

\FloatBarrier
\subsection{Trajectory reconstruction and interpolation in single-cell time courses}
\label{sec:real-holdout}
\label{sec:fitted-marginals}

Following the evaluation setting of WFR-FM \citep{peng2026wfrfm}, we use
two simulated gene-dynamics datasets (Simulation Gene and Dyngen
\citep{cannoodt2021dyngen}), a 1,000-dimensional Gaussian benchmark, and
four real single-cell time courses covering EMT \citep{cook2020context}, EB
development \citep{moon2019visualizing}, CITE-seq
\citep{lance2022multimodal}, and mouse hematopoiesis
\citep{weinreb2020lineage}.  Tables average \(W_1\) over the relevant
times, while dataset sizes and trajectory plots appear in
Appendices~\ref{app:dataset-coverage} and~\ref{app:trajectory-atlas}.

\paragraph{Matching observed population sequences.}
We fit one generator to every snapshot and generate from its initial
population, testing whether the learned drift, exit rate, and entry model
reproduce the observed sequence.

\begin{table}[t]
  \caption{Mean population-matching \(W_1\) across observed times.
  SF-EUOT (EOT), UDSB-F, and \method{} report mean\(\pm\)sample SD over
  five trained models, while WFR-FM and USB values are taken from their papers.
  Appendix \ref{app:population-matching-protocol} gives implementation
  details.}
  \label{tab:population-matching}
  \centering
  \footnotesize
  \renewcommand{\arraystretch}{1.08}
  \setlength{\tabcolsep}{3pt}
  \begin{tabular*}{\linewidth}{@{\extracolsep{\fill}}lccccc@{}}
    \toprule
    Dataset & WFR-FM & USB & \shortstack{SF-EUOT\\(EOT)} &
      UDSB-F & \method{} SDE \\
    \midrule
    \rawtableinput{tables/fitted_marginal_macro_rows.tex}
    \bottomrule
  \end{tabular*}
\end{table}

\method{} achieves the lowest mean \(W_1\) across all four real single-cell time courses and two of the three synthetic benchmarks, trailing USB only on the low-dimensional
Dyngen simulation.  The gains on all four real datasets
show that the analytic construction remains effective after neural
regression and stochastic generation in realistic state spaces
(Appendix~\ref{app:fitted-marginal-results}).

\paragraph{Interpolating an unobserved time point.}
We next omit one interior snapshot, refit the generator on the remaining
observations, and propagate the initial population continuously to the
omitted time with no observation reset, so each prediction tests the
learned dynamics between observations
(Appendix~\ref{app:real-holdout-protocol}).  The four real datasets provide
eight such predictions.  \method{} is best on EMT, EB, and Mouse.  On EB, mean
\(W_1\) falls from \(10.175\) for WFR-FM, the strongest baseline, to
\(9.770\), with gains at all three omitted times.
\rev{Appendix~\ref{app:real-sensitivity} explains the role of the
reference parameters in Table~\ref{tab:holdout-reference-settings} and
shows that replacing them with one shared default or a rule computed from the training snapshots raises population-matching \(W_1\) by at most \(7.1\%\) and holdout
\(W_1\) by at most \(6.0\%\).}

\subsection{Drug-induced expansion and contraction dynamics}
\label{sec:drug}

The 4i assay follows melanoma cells after combined EGFR/MEK inhibition,
during which the population first expands and then contracts while its
molecular state changes \citep{bunne2021learning,pariset2023unbalanced}.
All methods are trained on the same three snapshots and generate
continuously from 8 to 48 hours without resetting to the observed 24-hour
population, so errors made during expansion propagate into the contraction
phase.  \method{} models expansion through entry and contraction through
exit within one generator and attains the lowest \(W_1\) on held-out
cells at both 24 and 48 hours.  Appendix~\ref{app:drug-protocol}
reports the protocol and quantitative comparison.

\section{Discussion}
\label{sec:discussion}

By introducing a cemetery state under a killed-diffusion reference, \method{}
reduces path-space unbalanced SB inference to an endpoint projection paired
with analytic conditional bridge laws.  These closed-form components supply
simulation-free regression targets for the drift and exit rate, while a
generative flow models entry locations, recovering the optimal bridge in the
population limit.
Experiments verify recovery of hidden states, population size, and events
against independently solved benchmarks, and demonstrate accurate population matching,
time-point interpolation, and drug-response modeling on single-cell time courses.

\textbf{Limitations.}
Anonymous snapshots identify only the net exchange between consecutive times,
so the inferred exits, entries, and their states and times depend on
the chosen reference. Parent--offspring relations cannot be identified, and
each interval is modeled as net loss or net gain instead of concurrent
birth and death. \rev{Locally balanced turnover, for example, leaves the
expected population density unchanged, and with equal masses the
bridge has no events.  When one region grows while another shrinks within an interval, models with a
signed local growth rate fit better (Appendix~\ref{app:bbm}).} Addressing these limitations would require lineage
information \citep{weinreb2020lineage} and a branching reference
\citep{baradat2021branching}\rev{, and a reference that allows both exit
and entry within one interval would remove the restriction to net loss or
gain}. \rev{For single-cell data the masses are relative
abundances of profiled cells, and exits and entries describe changes in
this profiled population (Appendix~\ref{app:biology}).} The reference parameters \(\nu\) and
\(\kappa\) encode prior assumptions about matching scale and event timing.
In the sensitivity study of
Appendix~\ref{app:simulation-gene-sensitivity}, \(\nu\) leaves the mass
curves unchanged and has a modest effect on state interpolation, while a
large \(\kappa\) degrades both. \rev{On the real data the effect of \(\kappa\)
grows with the interval mass ratio (Appendix~\ref{app:real-sensitivity}).} \rev{The closed forms extend to
linear-drift references and to exit rates that depend only on time
(Appendices~\ref{app:linear-drift}--\ref{app:clock-requirements}), whereas
state-dependent exit rates generally break them.} Combining this killed-diffusion framework with lineage barcoding or metabolic RNA velocity measurements
\citep{lamanno2018rna} could further disentangle concurrent birth and death within a single interval.

\FloatBarrier
\subsection*{AI use statement}
The authors defined the research objective and core idea and supplied the
initial formulation and prototype code.  Part of the algorithmic design
used our agentic system \mbox{CellCompass}~\citep{zhang2026cellcompass},
and generative AI tools were used to assist with the mathematical
derivations, implementation refinement, numerical experiments, and
manuscript editing.
We have reviewed all AI-assisted work.  The derivations and proofs were checked
step by step by the authors, AI-assisted code was verified against
independent numerical checks, and all edited text was reviewed for
correctness.  The authors interpreted the results, verified the final
claims, and take responsibility for the final content of this work,
including text, claims, and artifacts produced with the aid of generative
AI.

\subsection*{Reproducibility statement}
Appendix~\ref{app:proofs} states all regularity assumptions and provides
complete derivations and proofs of the theoretical results.  Appendix~\ref{app:implementation} specifies the
common training and generation procedures, while Appendix~\ref{app:population-training-settings}
collects architectures, budgets, and reference clocks for each dataset.
Appendix~\ref{app:experiments} reports the evaluation metrics,
dataset coverage, training seeds, and
per-time results for every benchmark, including the sensitivity study over
the reference parameters.  The ground-truth benchmark protocol, including
its independent discrete solver and numerical checks, is described in
Appendix~\ref{app:formal-recovery}.

\bibliography{references}
\bibliographystyle{iclr2027_conference}

\appendix
\clearpage
\raggedbottom
\makeatletter
\setlength{\@fptop}{0pt}
\setlength{\@fpsep}{12pt}
\makeatother
\noindent\textbf{Guide to the appendices.}
Appendix~\ref{app:proofs} develops the proofs and technical extensions, and
Appendix~\ref{app:implementation} gives the common numerical implementation.
Appendix~\ref{app:experiments} follows the three experimental settings of Section~\ref{sec:experiments},
then collects sensitivity, cost, and trajectory visualizations.
Appendix~\ref{app:scope} compares models of growth and individual events
and explains the interpretation of population events.

\section{Proofs and technical details}
\label{app:proofs}

Appendix~\ref{app:abstract-usb-proof} proves the dynamic formulation,
cemetery encoding, and endpoint reduction in
Sections~\ref{sec:abstract-usb}--\ref{sec:cemetery-lift}.
Appendices~\ref{app:marked-reference}--\ref{app:proof-population-consistency}
give the marked reference, closed-form components, and recovery of the
learned generator.  Appendix~\ref{app:empirical-marked-consistency}
establishes convergence from empirical to population bridges, and
Appendix~\ref{app:event-clock-proof} extends the reference clocks.

\subsection{Notation}
\label{app:notation}

\begin{table}[!htbp]
  \caption{Notation shared by the main text and proofs.  Symbols used only
  within one proof are defined where they first appear.}
  \label{tab:notation}
  \centering
  \small
  \setlength{\tabcolsep}{5pt}
  \begin{tabular}{p{0.22\textwidth}p{0.69\textwidth}}
    \toprule
    Symbol & Meaning \\
    \midrule
    \(\state\), \(\bar\state=\state\cup\{\coffin\}\) & observed state space and its extension with a cemetery state \\
    \(\mu_t\), \(m_t\) & finite population measure at time \(t\) and its total mass \\
    \(g=|m_1-m_0|\) & population mass assigned to exit or time-reversed entry \\
    \(R\), \(P^*\) & killed reference path measure and its unbalanced SB projection \\
    \(X_t\), \(\tau\), \(\bar X_t\) & active state, exit time, and state set to the cemetery after exit \\
    \(R_{01}\), \(R^{xy}\), \(\pi^*\) & reference endpoint law, reference bridge conditioned on \((x,y)\), and optimal endpoint coupling \\
    \(\nu\), \(\kappa\), \(s_\kappa(t)\) & Brownian variance, reference event rate, and active fraction of an exit component \\
    \(Z_t\), \(b_t\), \(h_t\) & sampled bridge state and the drift and exit rate of its component \\
    \(b_\theta\), \(h_\theta\) & learned SDE drift and exit rate \\
    \(u_\phi\), \(T_\phi\) & auxiliary vector field for endpoint mixture centers and the map obtained by integrating this field \\
    \(\zeta\), \(\eta_t\), \(\rho_t\) & entry endpoint distribution used in the appendix, entry source, and active population measure \\
    \(\Delta_i\), \(D_i=\nu_i/\Delta_i\) & interval duration and diffusion coefficient in physical time \\
    \(v\), \(b^+\) & probability-flow velocity and forward SDE drift in the proofs (written as \(b\) in the main text) \\
    \bottomrule
  \end{tabular}
\end{table}

\subsection{The reference process, endpoint reduction, and dynamic formulation}
\label{app:abstract-usb-proof}

This subsection proves the endpoint KL decomposition and endpoint
reduction in Section~\ref{sec:cemetery-lift}, and the equivalent dynamic
formulation in Section~\ref{sec:abstract-usb}.  We use a general interval
\([0,T]\), where the main text takes \(T=1\).  Let the active endpoint masses be
\(m_0\geq m_T\), with total exit mass \(g=m_0-m_T\).  For a path measure
of mass \(m_0\), the notation \(\E_P F\) means \(\int F\,\dd P\),
or \(m_0\) times the expectation under its normalized path distribution.

\paragraph{Conditions for the continuous reference.}
The reference is a well-defined, nonexplosive Markov diffusion with
exits on \(\R^d\cup\{\coffin\}\), with finite initial mass \(m_0>0\).
Assume that the endpoint measures \(\mu_0\) and
\(\mu_T+g\delta_{\coffin}\) admit a coupling with finite KL relative to
\(R_{0T}\), as required in Proposition~\ref{prop:abstract-usb}.
Regular conditional path distributions are defined on the associated
standard Borel path space.  The empirical atomic problem uses the
marked reference in Appendix~\ref{app:marked-reference}, which does not
require an atomic terminal measure to be dominated by a Gaussian
transition density.

For the dynamic formulation, take \(\sigma\) to be invertible and
require the controls to satisfy the likelihood and integrability
conditions stated below.  Equivalence of the optimization problems
additionally requires the optimal bridge to admit the regular Doob
representation below, with controls satisfying those conditions.
The choice \(\sigma=\sqrt{\nu}I\) used in the remainder of the paper
satisfies the invertibility requirement.

\paragraph{Equivalence of cemetery encoding and the original path problem.}
Let \(J\) be the cemetery encoding in Section~\ref{sec:cemetery-lift}:
it retains the state path before exit and sets the state to \(\coffin\)
afterward.  Denote the encoded process by \(\bar X_t\) and the encoded
path measures by \(\bar P=J_\#P\) and \(\bar R=J_\#R\).
This encoding is invertible on histories over the observation interval.
If an encoded path reaches \(\coffin\), its first hitting time gives
\(\tau\), and its preceding segment gives the active path, while otherwise the
particle survives to \(T\).

Relative entropy is invariant under this encoding:
\[
 \KL(P\Vert R)=\KL(J_\#P\Vert J_\#R).
\]
The active marginal constraints in Equation~\ref{eq:abstract-usb}
correspond exactly to the complete endpoint measures
\(\mu_0,\mu_T+g\delta_{\coffin}\).  Setting \(T=1\) gives
Equation~\ref{eq:abstract-lift}, so the remaining calculations can be
carried out on the extended path space.

Specifically, let \(\bar P^*\) minimize the extended problem and set
\(P^*=(J^{-1})_\#\bar P^*\).  The endpoint correspondence ensures that
\(P^*\) matches the original active endpoint constraints.  If another
process \(Q\) matched those constraints with smaller
\(\KL(Q\Vert R)\), then \(J_\#Q\) would have smaller KL than
\(\bar P^*\) in the extended problem, contradicting optimality.
Thus \(P^*\) is optimal for the original problem.  The converse follows
by the same argument.

\paragraph{Endpoint KL decomposition and proof of endpoint reduction.}
Disintegrate \(P\) and \(R\) with respect to
\[
 \omega\longmapsto
 (\bar X_0(\omega),\bar X_T(\omega)).
\]
Write \(\pi=P_{0T}\) for the endpoint coupling and \(P^{xy},R^{xy}\)
for the conditional path laws.  Whenever \(P\ll R\), the
Radon--Nikodym derivative factorizes as
\[
 \frac{\dd P}{\dd R}(\omega)
 =
 \frac{\dd P_{0T}}{\dd R_{0T}}(\bar X_0,\bar X_T)
 \frac{\dd P^{\bar X_0\bar X_T}}
      {\dd R^{\bar X_0\bar X_T}}(\omega).
\]
Taking logarithms and integrating against \(P\) yields
\begin{equation}
 \KL(P\Vert R)
 =\KL(P_{0T}\Vert R_{0T})
 +\int_{\bar\state^2}
   \KL(P^{xy}\Vert R^{xy})\,P_{0T}(\dd x,\dd y).
 \label{eq:path-kl-chain-rule}
\end{equation}
Each path measure and its endpoint measure have the same total mass,
so the mass-correction terms in finite-measure KL are compatible with
this decomposition, and the conditional laws \(P^{xy},R^{xy}\) are
probability measures.  Taking \(T=1\) gives
Equation~\ref{eq:path-kl-chain-rule-main} \citep{leonard2014survey}.

For a fixed endpoint coupling, the conditional term is nonnegative and
vanishes precisely when \(P^{xy}=R^{xy}\) for \(\pi\)-almost every pair.
Every coupling with the prescribed marginals and finite KL can be
realized by mixing the reference conditional paths, with zero
conditional cost.  Minimizing path-space relative entropy is therefore
equivalent to minimizing \(\KL(\pi\Vert R_{0T})\) with the prescribed
endpoint marginals, as in Equation~\ref{eq:abstract-endpoint-projection}.

The set of couplings with fixed marginals is weakly compact, and
relative entropy is lower semicontinuous.  A coupling with finite KL
therefore ensures attainment, and strict convexity gives uniqueness.
The optimal path measure is
\begin{equation}
 P^*=\int_{\bar\state^2}R^{xy}\,\pi^*(\dd x,\dd y).
 \label{eq:abstract-component-mixture}
\end{equation}
Setting \(T=1\) gives
Equation~\ref{eq:abstract-component-mixture-main} and completes the
proof of Proposition~\ref{prop:abstract-usb}.

\paragraph{Derivation of the dynamic control cost.}
Let \(A_t=\mathbf 1_{\{t<\tau\}}\) and
\(N_t=\mathbf 1_{\{\tau\leq t\}}\).  The predictable intensity of \(N\)
is \(A_{t-}\lambda_0(t,X_{t-})\) under the reference and
\(A_{t-}\alpha_t\lambda_0(t,X_{t-})\) under \(P^{u,\alpha}\).
Thus the controlled drift is \(b_0+\sigma u\), and the exit rate is
\(h_t=\alpha_t\lambda_0(t,X_t)\), so \(P^{u,h}\) in the main text and
\(P^{u,\alpha}\) here are two parameterizations of the same path measure.
The controls are predictable and make the controlled diffusion and
exit process well defined.  Where \(\lambda_0=0\), finite cost requires
\(h_t=0\), and the value of \(\alpha_t\) there does not affect the process.

For positive exit rates, assume the likelihood process is a true
martingale and the control cost below is integrable.  The continuous
Girsanov formula and the likelihood formula for the exit counting
process give
\begin{align}
 \log\frac{\dd P^{u,\alpha}}{\dd R}
 &=\log\frac{\dd P_0}{\dd R_0}(X_0)
 +\int_0^T A_tu_t^\top\,\dd W_t^R
 -\frac12\int_0^T A_t\lVert u_t\rVert^2\,\dd t \notag\\
 &\quad+\int_0^T\log\alpha_t\,\dd N_t
 -\int_0^T A_t(\alpha_t-1)\lambda_0(t,X_t)\,\dd t.
 \label{eq:app-joint-likelihood}
\end{align}
Under the controlled process,
\[
 \dd W_t^R=\dd W_t^P+A_tu_t\,\dd t,
\]
and the compensator of \(N\) is
\[
 \int_0^t A_s\alpha_s\lambda_0(s,X_s)\,\dd s.
\]
Substitute the expression for \(W^R\) into
Equation~\ref{eq:app-joint-likelihood} and take expectations.
The Brownian integral has mean zero, while the mean of the counting
integral is determined by its compensator.  Hence
\begin{equation}
\begin{aligned}
 \KL(P^{u,\alpha}\Vert R)
 &=\KL(P_0\Vert R_0)\\
 &\quad+\E_{P^{u,\alpha}}\!
 \int_0^T A_t
 \left[
  \frac12\lVert u_t\rVert^2
  +\lambda_0(t,X_t)
   \{\alpha_t\log\alpha_t-\alpha_t+1\}
 \right]\dd t.
\end{aligned}
\label{eq:abstract-dynamic-action}
\end{equation}
In Equation~\ref{eq:abstract-usb}, \(P_0=R_0=\mu_0\), so the initial
relative-entropy term is zero.  Substituting
\(h_t=\alpha_t\lambda_0(t,X_t)\) gives
\[
 \lambda_0(t,X_t)\{\alpha_t\log\alpha_t-\alpha_t+1\}
 =h_t\log\frac{h_t}{\lambda_0(t,X_t)}-h_t+\lambda_0(t,X_t).
\]
Also, \(\int_0^T A_t(\cdot)\,\dd t=\int_0^{T\wedge\tau}(\cdot)\,\dd t\).
With \(T=1\), Equation~\ref{eq:abstract-dynamic-action} is exactly the
control cost in Equation~\ref{eq:dynamic-kl-main}, subject to the same
terminal constraint \(P_T^{\mathrm{act}}=\mu_T\).

The case \(h=0\) follows by positive approximation and lower
semicontinuity, with \(0\log(0/\lambda)=0\).
When \(\lambda=0\), the relative-entropy rate is zero for \(h=0\) and
infinite for \(h>0\).  Standard Markovization replaces history-dependent
controls by their conditional means given the current active state,
preserving the active marginals and not increasing the convex quadratic
and Poisson costs, provided the corresponding process exists.

\paragraph{Doob transform and equivalence of the dynamic optimization.}
Write \(a=\sigma\sigma^\top\).  Suppose the optimal endpoint projection
has the path-space potential representation
\[
 \frac{\dd P^*}{\dd R}=f(\bar X_0)q_T(\bar X_T).
\]
Here \(q_T\) denotes the terminal potential, with \(g\) reserved for
total exit mass.  Assume the backward potential is finite, strictly
positive on the active interior support, and \(C^{1,2}\) there.
Define
\[
 q_t(x)=\E_R[q_T(\bar X_T)\mid\bar X_t=x]
 =\begin{cases}
   \phi_t(x),&x\in\R^d,\\
   \psi_t,&x=\coffin.
  \end{cases}
\]
Because the cemetery is absorbing, \(\dot\psi_t=0\), while
\[
 \partial_t\phi+b_0\!\cdot\!\nabla\phi
 +\frac12a:\nabla^2\phi
 -\lambda_0(t,x)\phi+\lambda_0(t,x)\psi=0.
\]
For a test function \(F\), the time-dependent \(q_t\)-transform of the
reference generator is
\[
\begin{aligned}
 L_t^{P^*}F(x)
 &=\left(b_0+a\nabla\log\phi_t\right)\!\cdot\!\nabla F
   +\frac12a:\nabla^2F\\
 &\quad+\lambda_0(t,x)\frac{\psi_t}{\phi_t(x)}
   [F(\coffin)-F(x)].
\end{aligned}
\]
Its drift and exit rate are
\begin{equation}
 b^*(t,x)=b_0(t,x)+a(t,x)\nabla\log\phi_t(x),
 \qquad
 h^*(t,x)=\lambda_0(t,x)\frac{\psi_t}{\phi_t(x)}.
 \label{eq:abstract-optimal-controls}
\end{equation}
Set
\[
 u_t^*=\sigma(t,X_t)^\top\nabla\log\phi_t(X_t).
\]
Then \(b_0+\sigma u^*=b^*\), and \(h^*\) is the rate in
Equation~\ref{eq:abstract-optimal-controls}.  When these controls satisfy
the likelihood and integrability conditions above, the optimal bridge
belongs to the controlled class in the main text:
\(P^*=P^{u^*,h^*}\).

Every admissible control \((u,h)\) satisfying the endpoint constraints
defines a path measure in Equation~\ref{eq:abstract-usb}, and
Equation~\ref{eq:abstract-dynamic-action} equates its dynamic cost with
path-space relative entropy.  Thus the infimum of the dynamic problem
is at least the minimum of the path problem.  Conversely,
\((u^*,h^*)\) attains the path-space minimum and gives the reverse
inequality.  The two problems have the same optimal value and, by
uniqueness of the path-space optimizer, describe the same optimal
process.  This proves the equivalent dynamic formulation in
Section~\ref{sec:abstract-usb}.

These conditions specify the entropy and interior regularity
assumptions used in the calculation.  Identifying a generated process
uniquely from its learned coefficients additionally uses the
uniqueness-in-law conditions in
Appendix~\ref{app:proof-population-consistency}.

\subsection{Finite-measure KL and the marked reference}
\label{app:marked-reference}

For finite measures \(P\ll K\) on a common measurable space, we use
\begin{equation}
  \KL(P\Vert K)=
  \int\log\!\left(\frac{\dd P}{\dd K}\right)\dd P
  -P(\Omega)+K(\Omega).
  \label{eq:app-generalized-kl}
\end{equation}
Suppose a marked space is a disjoint union of components and
\(R=\sum_c k_cR^c\), \(P=\sum_cp_cP^c\), where \(R^c,P^c\) are probability
laws.  Direct substitution into \eqref{eq:app-generalized-kl} gives
\begin{equation}
  \KL(P\Vert R)=
  \sum_c\left[p_c\log\frac{p_c}{k_c}-p_c+k_c\right]
  +\sum_cp_c\KL(P^c\Vert R^c).
  \label{eq:app-kl-disintegration}
\end{equation}
The identity extends to zero component mass by lower semicontinuity.  Thus,
after endpoint component masses are chosen, each conditional path law is
uniquely minimized by its reference conditional law.

The empirical construction uses labeled atoms.  Coincident coordinates still
define distinct endpoint labels.  With Brownian bridge laws
\(\mathsf B_i^{ab}\), conditioned exit laws
\(\mathsf D_i^{a\coffin}\), and the idle cemetery path
\(\mathsf I^\coffin\), define the active and full marked references by
\begin{equation}
  \mathsf R_{i,\rm act}^{\rm disc}
  =\sum_{a,b}\widetilde K_{i,\rm act}^{ab}\mathsf B_i^{ab}
  +\sum_a\widetilde K_{i,\rm act}^{a\coffin}
     \mathsf D_i^{a\coffin},\qquad
  \overline{\mathsf R}_i^{\rm disc}
  =\mathsf R_{i,\rm act}^{\rm disc}+\mathsf I^\coffin .
  \label{eq:app-marked-reference}
\end{equation}
Forgetting labels maps it to the usual state-space interpolation.

Equation~\ref{eq:killed-kernel} is the dominated continuous reference with
\(\mu_0\) as its initial measure.  The same disintegration applies whenever
the prescribed coupling is dominated by this endpoint law.  Its live--live
endpoint density contains a factor for the source row, Gaussian normalization, and the survival factor
\(e^{-\kappa_i}\).  Multiplying a reference kernel by a positive row factor
and a positive column factor changes generalized KL by a constant on a set
of fixed marginals.  Hence the normalized continuous version of
\eqref{eq:abstract-endpoint-projection} is precisely the endpoint projection induced by
\eqref{eq:abstract-usb} through the lift \eqref{eq:abstract-lift}.

For empirical terminal atoms, the disjoint endpoint labels in
\eqref{eq:app-marked-reference} define the exact finite entropy problem
directly.  The dominated density formulation above is its continuous
population counterpart, and Proposition~\ref{prop:empirical-marked-consistency}
connects the active empirical bridges to that continuous projection as the
endpoint measures converge.

\subsection{Endpoint kernel of the killed Brownian reference}
\label{app:killed-kernel-derivation}

Condition on an active initial state \(X_0=x\).  The Brownian increment and
the exponential exit clock are independent.  For any measurable active set
\(A\subset\R^d\), survival to time one and the Gaussian transition therefore
give
\begin{equation}
 R_1(A\mid x)
 =\Pr(\tau>1)\Pr(x+\sqrt{\nu}W_1\in A)
 =e^{-\kappa}\int_A\varphi_\nu(y-x)\dd y.
 \label{eq:app-live-endpoint-kernel}
\end{equation}
All paths that exit by time one have the cemetery endpoint, so
\begin{equation}
 R_1(\{\coffin\}\mid x)=\Pr(\tau\leq1)=1-e^{-\kappa}.
 \label{eq:app-cemetery-endpoint-kernel}
\end{equation}
Multiplying these two conditional probabilities by the initial measure
\(\mu_0(\dd x)\) and placing them on their disjoint endpoint supports gives
Equation~\ref{eq:killed-kernel}.

\subsection{Bridge decomposition and cemetery slack (Theorem~\ref{thm:reduced-lift})}
\label{app:proof-reduced-lift}

We prove Theorem~\ref{thm:reduced-lift} together with an auxiliary invariance property showing that
adding the same cemetery mass at both endpoints introduces only paths that
remain at the cemetery, leaving the optimal active coupling and its
motion and exit process unchanged.

\begin{proof}
Orient time from larger to smaller mass.  The reference has no
paths from \(\coffin\) to a live state, so absolute continuity forces that block of
every feasible coupling with finite KL to vanish.  The cemetery source row has
mass \(s\), and therefore all of it must remain on the idle cemetery path.
The target cemetery column has mass \(g+s\), so its remaining live
contribution is
\begin{equation}
  (g+s)-s=g.
  \label{eq:app-forced-event-mass}
\end{equation}
Removing the forced idle atom leaves precisely the feasible set of
\eqref{eq:abstract-endpoint-projection}.  Generalized KL is additive on the
two disjoint blocks, and the idle contribution is constant.  Thus the active
coupling is unchanged and the added slack follows only the idle cemetery path.

The active feasible polytope is nonempty and compact.  Every allowed reference
entry is positive, so generalized KL is lower semicontinuous and strictly
convex on that support.  The active minimizer therefore exists and is unique.

Now disintegrate any active marked path measure \(P\) by its endpoint marks,
and write
\(\mathsf R_i^{ab}=\mathsf B_i^{ab}\) and
\(\mathsf R_i^{a\coffin}=\mathsf D_i^{a\coffin}\).
\Eqref{eq:app-kl-disintegration} becomes
\begin{equation}
  \KL(P\Vert\mathsf R_{i,\rm act}^{\rm disc})
  =\KL(\pi\Vert\widetilde K_{i,\rm act})
   +\sum_{u,v}\pi_{uv}
      \KL(P^{uv}\Vert\mathsf R_i^{uv}).
  \label{eq:app-path-chain-rule}
\end{equation}
For fixed \(\pi\), every conditional term is uniquely minimized by the
corresponding reference law, and the first term is uniquely minimized by
\(\pi_i^*\).  This proves the bridge decomposition in
Theorem~\ref{thm:reduced-lift}.

Finally, encode a killed component by its path before exit, exit time, and
endpoint marks.  Time reversal \(\Theta\) is a measurable bijection and
relative entropy is invariant under such a bijection:
\begin{equation}
  \KL(\Theta_\#P\Vert\Theta_\#R)=\KL(P\Vert R).
\end{equation}
It transposes the endpoint plan and turns exit into entry.
Applying it to both the optimizer in the loss orientation and its killed reference shows
that \(\Theta_\#\mathsf P_i^*\) is the unique projection relative to
\(\Theta_\#\mathsf R_{i,\rm act}^{\rm disc}\), which proves the
statement for a growing interval.
\end{proof}

\subsection{Conditional component laws and training targets}
\label{app:conditional-law-derivations}

We first give the components' active masses, then derive the state
distributions and local coefficients for the survival, exit, and entry
components in Section~\ref{sec:population-recovery}.

\paragraph{Total active mass.}
On the normalized interval \([0,1]\), orient time in the direction of
mass loss and write \(m_0\geq m_1\), with \(g=m_0-m_1\).
Survival components contribute mass \(m_1\) at every time, while exit
components remain active with fraction \(s_\kappa(t)\) from
Equation~\ref{eq:component-mass-factors}.  The total active masses of the
exit bridge and its time-reversed entry bridge are therefore
\begin{equation}
 m_{\mathrm{exit}}(t)=m_1+g s_\kappa(t),
 \qquad
 m_{\mathrm{entry}}(t)=m_1+g s_\kappa(1-t).
 \label{eq:mass-curves}
\end{equation}
For the instantaneous coefficients below, write
\(S_\kappa(r)=s_\kappa(r)\) and \(A_\kappa(r)=s_\kappa(1-r)\).

To obtain the SDE drift, we first find a velocity field that reproduces
each component's Gaussian marginals, then add the diffusion correction.
All derivatives below are in physical time, so
\(\partial_tr=1/\Delta_i\).  If
\(p_t=\Normal(m_t,q_tI_d)\), then
\begin{equation}
  u_t(z)=\dot m_t+\frac{\dot q_t}{2q_t}(z-m_t)
  \label{eq:app-gaussian-flow}
\end{equation}
satisfies \(\partial_tp_t+\nabla\!\cdot(p_tu_t)=0\).  This follows by
differentiating the Gaussian log density and matching its constant and
quadratic terms.

\paragraph{Survival (live--live).}
Gaussian conditioning gives
\[
  X_r\mid(X_0=x,X_1=y)
  \sim\Normal(m_r^{xy},\nu_i r(1-r)I_d).
\]
Here \(m_r^{xy}=(1-r)x+ry\).  For \(0<\varepsilon<1-r\), Gaussian
transition densities and Bayes' rule give the Brownian bridge transition
law
\[
 R^{xy}(X_{r+\varepsilon}\in\dd z'\mid X_r=z)
 =
 \frac{\varphi_{\nu_i\varepsilon}(z'-z)
       \varphi_{\nu_i(1-r-\varepsilon)}(y-z')}
      {\varphi_{\nu_i(1-r)}(y-z)}\,\dd z'.
\]
Its conditional mean is \(z+\varepsilon(y-z)/(1-r)\).
Dividing the mean increment by the physical time increment
\(\Delta_i\varepsilon\) gives
\[
 b_t^{+,xy}(z)
 =\frac{y-z}{\Delta_i(1-r)}
 =\frac{\nu_i}{\Delta_i}
   \nabla_z\log\varphi_{\nu_i(1-r)}(y-z).
\]
For \(\Delta_i=1\), this is the survival drift in
Section~\ref{sec:population-recovery}.  The component is conditioned on
survival to the endpoint, so its exit rate is zero.
Substituting the same Gaussian marginal into
Equation~\ref{eq:app-gaussian-flow} gives the probability-flow velocity
\[
 v^{xy}_t(z)=\frac{y-x}{\Delta_i}
 +\frac{1-2r}{2r(1-r)\Delta_i}(z-m_r^{xy}).
\]
Its score is
\(-[z-m_r^{xy}]/[\nu_i r(1-r)]\), and therefore
\begin{align}
  v^{xy}+\frac{D_i}{2}\nabla\log\rho^{xy}
  &=\frac{y-x}{\Delta_i}
    -\frac{z-m_r^{xy}}{(1-r)\Delta_i}
   =\frac{y-z}{(1-r)\Delta_i}.
  \label{eq:app-ll-forward}
\end{align}

\paragraph{Exit (live--cemetery).}
Conditioning the independent exponential clock on death by normalized time
one gives
\begin{equation}
  \Pr(\tau>r\mid\tau\leq1)
  =S_{\kappa_i}(r).
\end{equation}
Conditional on being live at \(r\), the spatial law is
\(\Normal(x,\nu_i rI_d)\).  \Eqref{eq:app-gaussian-flow} gives
\(v=(z-x)/(2r\Delta_i)\), while its score contribution
\((D_i/2)\nabla\log\rho\) is the negative of this quantity.  Hence
\(b^+=0\).

For \(0<\varepsilon<1-r\), conditional on being active at \(r\) and
exiting by the endpoint, the probability of exit in the next short
interval is
\[
 \Pr(r<\tau\leq r+\varepsilon\mid r<\tau\leq1)
 =
 \frac{e^{-\kappa_i r}-e^{-\kappa_i(r+\varepsilon)}}
      {e^{-\kappa_i r}-e^{-\kappa_i}}.
\]
Dividing by \(\Delta_i\varepsilon\) and sending
\(\varepsilon\downarrow0\) gives the conditional exit rate, equivalently
the negative logarithmic derivative of the active fraction:
\begin{equation}
  -\partial_t\log S_{\kappa_i}(r)
  =\frac{\kappa_i}
  {\Delta_i[1-e^{-\kappa_i(1-r)}]}=h_t^{x\coffin}.
  \label{eq:app-death-hazard}
\end{equation}
Although the conditional exit rate diverges at the terminal boundary, its
exit flux remains regular:
\begin{equation}
  S_{\kappa_i}(r)h_t^{x\coffin}
  =\frac{\kappa_i e^{-\kappa_i r}}
   {\Delta_i(1-e^{-\kappa_i})},\qquad
  \int_{t_i}^{t_{i+1}}S_{\kappa_i}(r)h_t^{x\coffin}\dd t=1.
  \label{eq:app-finite-death-flux}
\end{equation}

\paragraph{Entry (cemetery--live).}
Time reversal replaces \(r\) with \(1-r\).  The live finite measure is
\(A_{\kappa_i}(r)\Normal(y,\nu_i(1-r)I_d)\).
\Eqref{eq:app-gaussian-flow} gives
\(v_t(z)=-(z-y)/[2(1-r)\Delta_i]\), and adding half the diffusion times
the score gives
\[
  b^+=-\frac{z-y}{(1-r)\Delta_i}.
\]
The component stays active after entry, so \(h_t^{\coffin y}=0\).
Its entry events are described by a source with temporal rate
\begin{equation}
  \partial_tA_{\kappa_i}(r)
  =\frac{\kappa_i e^{-\kappa_i(1-r)}}
   {\Delta_i(1-e^{-\kappa_i})}
  =a_{\kappa_i}(r).
  \label{eq:app-birth-rate}
\end{equation}
Weighting these entry contributions by the endpoint coupling gives
the source in Equation~\ref{eq:entry-source}.
The total entering mass of a unit entry component is one.

Each component consequently satisfies
\begin{equation}
  \partial_t\rho_t^c+\nabla\!\cdot(\rho_t^cv_t^c)
  =-h_t^c\rho_t^c+\eta_t^c.
  \label{eq:app-component-continuity}
\end{equation}

\subsection{Recovery of the entry source (Proposition~\ref{prop:entrance-amortization})}
\label{app:proof-entry-flow}

We prove Proposition~\ref{prop:entrance-amortization} and analyze the
training objective for the entry map \(T_\phi\) in
Section~\ref{sec:entrance}.
For one normalized growing interval, define the endpoint distribution
of the entry components by
\[
 \zeta=\frac1g\sum_y\pi^*_{\coffin y}\delta_y,
 \qquad g=\sum_y\pi^*_{\coffin y}>0.
\]
We use \(\zeta\) only in the auxiliary-flow and error calculations below, whereas
the main text uses the endpoint weights from the coupling directly.
For continuous endpoints, replace the sums by integrals against the
entry endpoint measure.

\paragraph{Recovery of the entry source.}
If \(U\sim\Normal(0,I)\) and \(\mathcal L(T_\phi(U))=\zeta\), then,
conditional on \(T_\phi(U)=y\),
Equation~\ref{eq:generated-birth-source} gives density
\(\varphi_{\nu(1-t)}(z-y)\).
Integrating over endpoints and multiplying by the total entry rate
gives
\[
\begin{aligned}
 &g\,\frac{\dd}{\dd t}s_\kappa(1-t)
   \int\varphi_{\nu(1-t)}(z-y)\,\zeta(\dd y)\\
 &\qquad=
   \frac{\dd}{\dd t}s_\kappa(1-t)
   \sum_y\pi^*_{\coffin y}\varphi_{\nu(1-t)}(z-y)
   =\eta_t(z).
\end{aligned}
\]
This is the source in Equation~\ref{eq:entry-source}, proving
Proposition~\ref{prop:entrance-amortization}.

\paragraph{Training the entry map.}
Draw \(U\sim\Normal(0,I)\), \(Y\sim\zeta\), and auxiliary flow time
\(s\sim\mathrm{Unif}[0,1]\) independently, and let
\(Z_s=(1-s)U+sY\).  Here \(s\) is not the physical time \(t\) of the
population process.  The training objective is
\begin{equation}
 \mathcal L_{\mathrm{entry}}(\phi)
 =\E_{s,U,Y}\left\lVert u_\phi(s,Z_s)-(Y-U)\right\rVert^2.
 \label{eq:entrance-flow-matching}
\end{equation}
At generation time, integrating \(u_\phi\) from \(s=0\) to \(s=1\),
starting from \(U\), gives the endpoint mixture center \(T_\phi(U)\).
Let \(\chi_s=\mathcal L(Z_s)\).  The population minimizer of squared
error is
\[
 u^*(s,z)=\E[Y-U\mid Z_s=z].
\]
For every compactly supported smooth function \(f\),
\[
 \frac{\dd}{\dd s}\E f(Z_s)
 =\int\nabla f(z)^\top u^*(s,z)\,\chi_s(\dd z).
\]
This is the weak continuity equation
\(\partial_s\chi_s+\nabla\!\cdot(\chi_su^*)=0\).
It starts from \(\Normal(0,I_d)\) and converges weakly to \(\zeta\) as
\(s\uparrow1\).  For an atomic empirical target, the endpoint is
understood in this weak limit.

\paragraph{Source error.}
Let the learned endpoint distribution be
\(\widehat\zeta=\mathcal L(T_\phi(U))\).
Couple \(Y\sim\zeta\) and \(\widehat Y\sim\widehat\zeta\), and add the
same independent Gaussian noise to both.  Their expected distance is
unchanged.  Taking the infimum over couplings gives
\begin{equation}
 W_1\!\left(
 \frac{\eta_t}{\eta_t(\R^d)},
 \frac{\widehat\eta_t}{\widehat\eta_t(\R^d)}
 \right)
 \leq W_1(\zeta,\widehat\zeta).
 \label{eq:source-flow-w1-bound}
\end{equation}
The common scalar source mass cancels after normalization.

\subsection{Recovery of the optimal path law (Theorem~\ref{thm:population-consistency})}
\label{app:proof-population-consistency}

\paragraph{Assumptions and endpoint interpretation.}
For the continuous formulation, use the conditions for endpoint projection and
the interior Doob transform in Appendix~\ref{app:abstract-usb-proof}, with
\(\nu,\kappa>0\).  At each interior time, assume the drift labels have
finite second moments and the conditional Poisson loss is finite for an
admissible rate.  Regression optimality is understood pointwise for almost
every \(t\in(0,1)\), or on compact interior time intervals, and does not
require uniformly bounded component targets at the endpoints.

For the resulting drift and exit rate, assume the killed-diffusion
martingale problem, initialized by \(\mu_0/m_0\), has a unique nonexplosive
law on every \([0,1-\epsilon]\), \(\epsilon\in(0,1)\), and admits terminal path
limits in the augmented state space.  On a gaining interval, require the
analogous uniqueness with its prescribed initial population, entry clock,
and kernel for entering states.  These are the uniqueness-in-law conditions used
in Theorem~\ref{thm:population-consistency}, while existence is supplied by the
reference bridge construction.  Local spatial Lipschitz regularity and
linear growth of the drift, together with a locally bounded nonnegative
exit rate, provide sufficient local conditions in the loss orientation.
Terminal coefficients may be singular, with endpoint measures recovered
as weak traces as shown below.

For a finite endpoint plan, interpret the target as the marked bridge of
Appendix~\ref{app:marked-reference} after forgetting endpoint labels.
Its potentials, which are sums of Gaussians, give the corresponding interior construction, while
convergence to the population bridge is addressed separately in
Appendix~\ref{app:empirical-marked-consistency}.

\begin{proof}
\emph{Regression optima.}
Fix an interval and an interior time.  With weights from the endpoint plan
\(\omega_c\),
\[
  \rho_t(z)=\sum_c\omega_c\rho_t^c(z),\qquad
  \Pr(C_t=c\mid Z_t=z)
  =\frac{\omega_c\rho_t^c(z)}{\rho_t(z)}.
\]
The targets in the main text are \(b_t=b_t^{+,C_t}(Z_t)\) and
\(h_t=h_t^{C_t}\).  Squared loss is pointwise minimized by conditional
expectation, which proves the identity for the forward drift.  Define \(v^*\) as the conditional mean of the
analytic component velocities.  For every component, its forward SDE drift
and probability-flow velocity satisfy
\begin{equation}
 b^{+,c}=v^c+\frac{D_i}{2}\nabla\log\rho_t^c .
 \label{eq:conditional-forward-drift}
\end{equation}
Therefore,
\begin{align}
 b^{+*}-v^*
 &=\frac{D_i}{2}\sum_c
   \frac{\omega_c\rho_t^c}{\rho_t}\nabla\log\rho_t^c
 =\frac{D_i}{2}\frac{\sum_c\omega_c\nabla\rho_t^c}{\rho_t}
 =\frac{D_i}{2}\nabla\log\rho_t .
 \label{eq:app-mixture-forward-drift}
\end{align}

For the exit-rate term, let \(U=h_t^{C_t}\).  At fixed \((t,z)\), the
derivative of
\(\E[U\log(U/b)-U+b\mid Z_t=z]\) with respect to \(b>0\) is
\[
  \frac{\partial}{\partial b}
  \E[\ell_{\rm P}(U,b)\mid Z_t=z]
  =1-\frac{\E[U\mid Z_t=z]}{b}.
\]
Its minimizer is the conditional mean, with zero obtained in the closure.

\emph{Identification of the path law.}
We next identify these conditional means with the unbalanced SB optimizer and
thereby recover its full path law.  By
\eqref{eq:abstract-component-mixture-main}, drawing a component using the
mass assigned by the endpoint plan and then drawing its conditioned path is exactly a
disintegration of \(P^*\).  Moreover, the endpoint entropy projection has
\(\dd P^*/\dd R=f(\bar X_0)q_T(\bar X_T)\).  It is therefore the Markov Doob transform of
the killed reference, with drift and exit intensity given by
\eqref{eq:abstract-optimal-controls}.  The predictable characteristics of
this same path mixture are the conditional component averages derived above.
Consequently \(b^{+*}\) and \(h^*\) are the local characteristics of
\(P^*\).  By the uniqueness-in-law condition stated above, the killed
diffusion with these characteristics in the loss orientation has path law
\(P^*\).

In the empirical case, the same argument uses the marked reference instead of
absolute continuity with respect to the continuous Brownian reference.
On rows and columns with positive mass, the optimal finite plan is a positive
row--column scaling of the endpoint kernel.  Combining its column weights with the Gaussian transition kernels
and cemetery transition probability gives the backward potential as a finite sum.
The Gaussian transition identity then identifies the unmarked mixture as
its Markov transform at interior times, with the same conditional drift
and exit-rate formulas.  The uniqueness argument therefore applies to this
empirical target as well.

\emph{Population evolution.}
Multiplying \eqref{eq:app-component-continuity} by \(\omega_c\) and summing
gives
\[
 \sum_c\omega_c\rho_t^cv_t^c=\rho_tv^*,\qquad
 \sum_c\omega_c\rho_t^ch_t^c=\rho_th^*.
\]
The weighted entry terms sum to the source \(\eta_t\) in
Equation~\ref{eq:entry-source}, hence
\begin{equation}
 \partial_t\rho_t+\nabla\!\cdot(\rho_tv^*)
 =-h^*\rho_t+\eta_t.
 \label{eq:app-mixture-continuity}
\end{equation}

In physical time, the diffusion coefficient is
\(D_i=\nu_i/\Delta_i\), and setting \(\Delta_i=1\) gives the normalized time
notation in the main text.  The spatial Fokker--Planck operator of
the SDE in \eqref{eq:sde-inference} is
\begin{align}
 -\nabla\!\cdot(\rho_tb^{+*})
 +\frac{D_i}{2}\Delta\rho_t
 =-\nabla\!\cdot(\rho_tv^*).
 \label{eq:app-fp-cancellation}
\end{align}
Combining this identity with \eqref{eq:app-mixture-continuity}, and returning
to normalized time, gives \eqref{eq:fixed-pool-continuity}.

\emph{Time reversal and endpoint traces.}
For a gaining interval, Proposition~\ref{prop:entrance-amortization}
reconstructs the exact entry source.  Time reversal of the
loss-oriented bridge then gives the same local drift, entry clock, and entry
state kernel as the construction in the main text.  Uniqueness again
identifies the resulting augmented path law with the time reversal of
\(P^*\).

Finally, live--live Gaussians converge weakly to their endpoint atoms,
\(s_\kappa(0)=1,s_\kappa(1)=0\), and therefore
\(s_\kappa(1-0)=0,s_\kappa(1-1)=1\).  The row and column constraints of the plan
therefore give the prescribed endpoint traces, and time reversal gives the
gaining case.
\end{proof}

\subsection{Consistency from empirical to population bridges}
\label{app:empirical-marked-consistency}

The preceding construction uses a finite endpoint plan in practice.  The
following result connects its marked bridge to the continuous population
problem as the empirical endpoints converge.

\begin{proposition}[Consistency of empirical marked bridges]
\label{prop:empirical-marked-consistency}
Let the live state space \(\mathsf X\subset\R^d\) be compact.  Suppose the
empirical endpoint measures of larger and smaller mass converge weakly to finite
measures on \(\mathsf X\), and their masses converge.  For fixed
\(\nu_i,\kappa_i>0\), the active solutions of
\eqref{eq:abstract-endpoint-projection} converge weakly to the unique continuous
endpoint entropy projection.  If \(\rho_{i,r}^{(n)}\) and \(\rho_{i,r}\)
are the corresponding active marginals of the bridge mixtures, then, for every
\(\epsilon\in(0,1/2)\),
\begin{equation}
 \sup_{r\in[\epsilon,1-\epsilon]}
 d_{\rm BL}\!\left(\rho_{i,r}^{(n)},\rho_{i,r}\right)\longrightarrow0.
 \label{eq:empirical-marked-marginal-convergence}
\end{equation}
The finite measures of exit or entry times and event locations converge
weakly as well.  When the limiting endpoints admit a coupling with finite KL
relative to the continuous reference, the limit is the killed-diffusion
unbalanced SB of Definition~\ref{def:usb}, with local dynamics given by
Equation~\ref{eq:abstract-optimal-controls} under the interior regularity
conditions above.  The gaining case follows by time
reversal.
\end{proposition}

\begin{proof}
We give the case of decreasing mass.  Let \(H_n,L_n\) be the empirical endpoint
masses and \(H,L\) their limits.  Write
\(\varsigma_n=\mu_n^L+(H_n-L_n)\delta_{\coffin}\) and
\(\varsigma=\mu^L+(H-L)\delta_{\coffin}\).  On the compact augmented
endpoint space, after deleting a zero-mass cemetery column if necessary,
\[
 k(x,y)=
 \begin{cases}
 e^{-\kappa_i}\varphi_{\nu_i}(y-x),&y\in\mathsf X,\\
 1-e^{-\kappa_i},&y=\coffin
 \end{cases}
\]
is continuous and bounded above and away from zero on its allowed support.
Multiplying rows or columns of the empirical reference by positive
weights changes generalized KL only by terms fixed by the marginals.
Consequently \eqref{eq:abstract-endpoint-projection} has the same optimizer as the
entropic projection with reference
\(k(x,y)\mu_n^H(\dd x)\varsigma_n(\dd y)\).

Fixing the additive gauge in its Schr\"odinger dual, the uniform upper and
lower bounds on \(k\) bound the dual potentials.  The Sinkhorn equations and
uniform continuity of \(k\) make their continuous extensions
equicontinuous.  Arzel\`a--Ascoli gives a uniformly convergent subsequence, and
weak endpoint convergence permits passage to the limit in both Sinkhorn
equations.  Every subsequential limit therefore solves the continuous
projection.  Strict convexity on the allowed support makes that solution
unique, so the complete sequence of plans satisfies
\(\pi_n^*\Rightarrow\pi^*\).

For \(r\in[\epsilon,1-\epsilon]\), the Gaussian live--live and
live--absorbing conditional kernels depend uniformly continuously, in
bounded--Lipschitz distance, on their endpoint labels and on \(r\).
Integrating them against the convergent plans proves
\eqref{eq:empirical-marked-marginal-convergence}.  Event measures are
continuous mixtures of the same plan and the fixed conditional clock, so
they converge as well.  Row--column gauge invariance identifies the limiting
plan with Proposition~\ref{prop:abstract-usb} whenever the limiting
endpoint problem admits a feasible coupling with finite KL.  Time reversal gives
the gaining case.
\end{proof}

\subsection{Extensions to time-dependent reference clocks}
\label{app:event-clock-proof}

We now allow the reference rate to vary over time without changing the endpoint
projection.  A rate that depends only on time changes event timing while leaving the
spatial reference unchanged, whereas a deterministic time change reparameterizes
both motion and events.  Their experimental settings are collected in
Appendix~\ref{app:population-training-settings}.

\paragraph{Event rates that depend only on time.}
Let \(c_i:[0,1]\to(0,\infty)\) satisfy
\(\int_0^1c_i(u)\dd u=1\), and write
\(C_i(r)=\int_0^r c_i(u)\dd u\).  The conditional exit density and
remaining live fraction are
\begin{equation}
 q^-_{\kappa_i,c_i}(r)
 =\frac{\kappa_i c_i(r)e^{-\kappa_i C_i(r)}}
        {1-e^{-\kappa_i}},\qquad
 S_{\kappa_i,c_i}(r)
 =\frac{e^{-\kappa_i C_i(r)}-e^{-\kappa_i}}
        {1-e^{-\kappa_i}},
 \label{eq:general-event-clock}
\end{equation}
with entry obtained by \(r\mapsto1-r\).  The Mouse experiment uses
\begin{equation}
 c_\beta(r)=
 \begin{cases}
 \beta e^{\beta r}/(e^\beta-1),&\beta\ne0,\\
 1,&\beta=0,
 \end{cases}
 \qquad
 C_\beta(r)=
 \begin{cases}
 (e^{\beta r}-1)/(e^\beta-1),&\beta\ne0,\\
 r,&\beta=0.
 \end{cases}
 \label{eq:exponential-event-clock}
\end{equation}

\begin{proposition}[Event rates that depend only on time]
\label{prop:time-only-event-clock}
Let \(c_i:[0,1]\to(0,\infty)\) be continuous with
\(\int_0^1c_i(u)\dd u=1\).  Replacing the constant rate in
\eqref{eq:reference-normalized} by the rate
\(\kappa_i c_i(r)\), independent of the Brownian motion, leaves the endpoint kernel and endpoint entropy
projection unchanged.  Conditional on an event before time one, the exit
density and remaining live fraction are those in
\eqref{eq:general-event-clock}.  Time reversal gives entrance, and the
conditional spatial Brownian bridges are unchanged.
\end{proposition}

\begin{proof}
The unconditioned survival function is
\(\exp[-\kappa_i C_i(r)]\), and \(C_i(1)=1\), so survival through the whole interval
is still \(e^{-\kappa_i}\).  Independence from Brownian motion gives the
same endpoint kernel.  Differentiating the survival function, conditioning
on an event by time one, and subtracting the terminal survivor gives
\eqref{eq:general-event-clock}.  Time reversal gives entrance.
\end{proof}

Once the reference clock is fixed, its conditional event law is selected by
KL.  Let
\(P_{\kappa,c}|_{[0,1]}=(1-e^{-\kappa})Q_{\kappa,c}^-\), where
\(Q_{\kappa,c}^-\) has density \(q_{\kappa,c}^-\).  For any event-time
probability \(Q\ll Q_{\kappa,c}^-\),
\begin{equation}
  \KL(Q\Vert P_{\kappa,c}|_{[0,1]})
  =\KL(Q\Vert Q_{\kappa,c}^-)
   -\log(1-e^{-\kappa})-e^{-\kappa}.
  \label{eq:app-clock-kl}
\end{equation}
The unique minimizer is therefore \(Q_{\kappa,c}^-\).  Multiplication by the
fixed mass \(g\) changes only fixed scalar terms.

\paragraph{Deterministic time changes.}
\begin{corollary}[Deterministic reference clocks]
\label{cor:deterministic-reference-clock}
Let \(\psi_i:[0,1]\to[0,1]\) be a continuously differentiable increasing
bijection with \(\psi_i'(r)>0\).  The time-changed reference
\(\widetilde X_r=X_{\psi_i(r)}\) has
\begin{equation}
 \dd\widetilde X_r
 =\sqrt{\nu_i\psi_i'(r)}\,\dd\widetilde W_r,\qquad
 \widetilde\lambda_i(r)=\kappa_i\psi_i'(r),
 \label{eq:time-changed-reference}
\end{equation}
and the same endpoint kernel.  Its unbalanced SB is the deterministic time
change of the base unbalanced SB.  Replace \(r\) by \(\psi_i(r)\) in every conditional law and
multiply drift, exit rate, source rate, and quadratic-variation rate by
\(\psi_i'(r)\).
\end{corollary}

\begin{proof}
The map \(\mathcal T_{\psi_i}:\omega\mapsto
[r\mapsto\omega(\psi_i(r))]\) is a measurable path-space bijection that
fixes both endpoints.  Standard deterministic time change gives
\eqref{eq:time-changed-reference}, and therefore the endpoint kernel is
unchanged.  Relative entropy is invariant under a measurable bijection:
\[
 \KL((\mathcal T_{\psi_i})_\#P\Vert
     (\mathcal T_{\psi_i})_\#R)=\KL(P\Vert R).
\]
Applying this equality to the bridge variational problem proves that the
optimizer is \((\mathcal T_{\psi_i})_\#P^*\).  The chain rule gives the
stated transformations of rates and quadratic variation.  Time reversal
gives the gaining case.
\end{proof}

\begin{revised}
\subsection{Static, multi-marginal, and iterative views of the endpoint projection}
\label{app:static-structure}

\paragraph{The static problem.}
By \eqref{eq:path-kl-chain-rule-main}, a path measure with endpoint
coupling \(\pi\) has relative entropy at least \(\KL(\pi\Vert R_{01})\),
with equality for the mixture \(\int R^{xy}\pi(\dd x,\dd y)\), so the
path problem is equivalent to the static problem
\eqref{eq:abstract-endpoint-projection} \citep{leonard2014survey}.  For a
Brownian reference without exit,
\(R_{01}(\dd x,\dd y)\propto\mu_0(\dd x)\exp(-\lVert x-y\rVert^2/2\nu)\,\dd y\),
and for couplings with fixed marginals
\[
 \KL(\pi\Vert R_{01})=\frac1{2\nu}\int\lVert x-y\rVert^2\,\pi(\dd x,\dd y)
 +\KL(\pi\Vert\mu_0\otimes\mu_1)+\text{const},
\]
which is entropic OT with regularization \(\nu\) (for atomic
\(\mu_1\), with the marked reference of Appendix~\ref{app:marked-reference}).  On
samples, Sinkhorn alternately rescales the rows and columns of the kernel
matrix to match the two marginals.  This is iterative proportional
fitting on the endpoint law, the static counterpart of the path-space IPF
of DSB, whose half-bridges alternately match the initial and terminal
marginals \citep{debortoli2021dsb}.  Sinkhorn therefore returns, to solver
tolerance, the static Schr\"odinger coupling of the retained empirical
endpoint measures, which is the object to which the forward--backward iterations
converge.  The killed kernel in Equation~\ref{eq:killed-kernel} multiplies the
Gaussian block by \(e^{-\kappa}\) and adds one cemetery column, and the
same scaling iterations apply.  The coupling is computed once for all
retained samples before neural training, and Appendix~\ref{app:ot-cap}
reports its cost.

\paragraph{Relation to iterative Markovian fitting.}
Bridge-matching solvers alternate a reciprocal projection, which mixes
reference bridges over the current endpoint coupling, with a Markovian
projection, which regresses the bridge drift on time and current state
\citep{shi2024dsbm,peluchetti2023diffusion}.
Theorem~\ref{thm:reduced-lift} gives the reciprocal projection for the
killed reference, and the regression of
Theorem~\ref{thm:population-consistency} is its Markovian projection,
including the exit rate.  Started from the exact static solution
\(\pi^*\), the reciprocal mixture is already \(P^*\) by
Proposition~\ref{prop:abstract-usb}, so a single Markovian projection
recovers the bridge and further iterations leave it unchanged.
\rev{\citet{shi2024dsbm} note the same for balanced bridges, showing that
DSBM converges in one iteration when started from the static SB coupling.}
Iteration is needed only when \(\pi^*\) is not computed directly.  When
the endpoint problem is too large for Sinkhorn, a forward--backward variant
could re-estimate the coupling from simulated killed paths with the same two projections.  We did not need it at the sample sizes considered here.

\paragraph{Several snapshots.}
Let \(t_0<\dots<t_K\) be observation times and let \(\bar\mu_k\) be the
cemetery-lifted snapshot at \(t_k\), so that all \(\bar\mu_k\) have the
same total mass.  For a path measure \(R\) on \([t_0,t_K]\) and \(k<K\),
let \(R^{(k)}\) be the law of \(R\) on \([t_k,t_{k+1}]\) restarted from
\(\bar\mu_k\), that is,
\(R^{(k)}=\int\bar\mu_k(\dd x)\,R(X_{[t_k,t_{k+1}]}\in\cdot\mid X_{t_k}=x)\).

\begin{proposition}[Markov references separate across intervals]
\label{prop:multi-marginal}
Let \(R\) be Markov on \(\bar\state\), and suppose that for each \(k<K\)
the two-marginal problem
\(\min\{\KL(Q\Vert R^{(k)}):Q_{t_k}=\bar\mu_k,\ Q_{t_{k+1}}=\bar\mu_{k+1}\}\)
has a solution \(Q_k^*\) with finite value.  Then the multi-marginal
problem
\[
 \min\{\KL(P\Vert R):P_{t_k}=\bar\mu_k,\ k=0,\dots,K\}
\]
has the unique solution \(P^*\) obtained by joining
\(Q_0^*,\dots,Q_{K-1}^*\) at the shared snapshots as a Markov process,
and
\(\KL(P^*\Vert R)=\KL(\bar\mu_0\Vert R_{t_0})+\sum_{k<K}\KL(Q_k^*\Vert R^{(k)})\).
\end{proposition}

\begin{proof}
Let \(P\) be feasible with finite relative entropy, and let
\(\mathcal F_k\) be the history up to \(t_k\).  The chain rule over the
segments \([t_k,t_{k+1}]\) gives
\[
 \KL(P\Vert R)=\KL(\bar\mu_0\Vert R_{t_0})
 +\sum_{k<K}\E_P\,\KL\big(P(X_{[t_k,t_{k+1}]}\in\cdot\mid\mathcal F_k)
 \,\big\Vert\,R(X_{[t_k,t_{k+1}]}\in\cdot\mid\mathcal F_k)\big).
\]
By the Markov property, \(R(\cdot\mid\mathcal F_k)=R(\cdot\mid X_{t_k})\).
For random elements \(U,V,W\),
\(\E\KL(P_{V\mid U,W}\Vert R_{V\mid W})
 =\E\KL(P_{V\mid W}\Vert R_{V\mid W})+I_P(U;V\mid W)\), so the \(k\)th
summand is at least
\(\E_P\KL(P(X_{[t_k,t_{k+1}]}\in\cdot\mid X_{t_k})\Vert R(X_{[t_k,t_{k+1}]}\in\cdot\mid X_{t_k}))
=\KL(P_{[t_k,t_{k+1}]}\Vert R^{(k)})\), where the equality uses
\(P_{t_k}=\bar\mu_k\), and the bound is attained exactly when the
past and the future of \(P\) are conditionally independent given
\(X_{t_k}\).  Each segment law \(P_{[t_k,t_{k+1}]}\) is feasible for the
\(k\)th two-marginal problem, hence
\(\KL(P\Vert R)\ge\KL(\bar\mu_0\Vert R_{t_0})+\sum_k\KL(Q_k^*\Vert R^{(k)})\).
The Markov concatenation of the \(Q_k^*\) is feasible, because consecutive
pieces share the marginal \(\bar\mu_{k+1}\), and attains the bound.
Uniqueness follows from strict convexity of relative entropy.
\end{proof}

When every interval loses mass, the killed Brownian reference is Markov on
\(\bar\state\) with an absorbing cemetery.  Then \(R^{(k)}\) is the killed
reference started from \(\mu_k\) together with the mass already at
\(\coffin\), and the \(k\)th problem is the per-interval projection of
Section~\ref{sec:cemetery-lift} with exit mass \(m_k-m_{k+1}\).  The
per-interval construction used by \method{} therefore solves the
multi-marginal problem for such sequences, and balanced sequences with a
Brownian reference are the case \(\kappa=0\).  On a growing interval the
reference is defined by time reversal from the final snapshot of that
interval (Section~\ref{sec:abstract-usb}), so for mixed sequences the
generator is the Markov concatenation of per-interval bridges.  References that are not Markov in
the observed state, such as the momentum references of
\citet{theodoropoulos2025momentum}, do not separate in this way, which is
why such methods condition bridges on several snapshots jointly.
\rev{IMF-based multi-marginal solvers \citep{gravier2026mmtsbm,park2025msbm}
refresh the coupling by simulating the learned process, and
\citet{park2025msbm} likewise assemble the solution from local bridges on
each interval.}

\subsection{Linear-drift references}
\label{app:linear-drift}

The closed forms of Theorem~\ref{thm:reduced-lift} use only that the
reference motion is Gaussian and that its exit clock is independent of
it.  Both hold for the linear-drift reference
\[
 \dd X_t=(AX_t+a)\,\dd t+\sqrt{\nu}\,\dd W_t\quad(t<\tau),
\]
with \(A\in\R^{d\times d}\), \(a\in\R^d\), and an exit time independent
of \(W\) whose rate depends only on time
(Appendix~\ref{app:clock-requirements}).  Write \(\Phi_t=e^{At}\),
\(m_t(x)=\Phi_tx+\int_0^t\Phi_s\,\dd s\,a\), and
\(\Sigma_t=\nu\int_0^t\Phi_s\Phi_s^\top\dd s\), so that an active
particle started at \(x\) satisfies \(X_t\sim\Normal(m_t(x),\Sigma_t)\).
For the constant rate \(\kappa\):

\begin{itemize}[leftmargin=*,itemsep=1pt,topsep=2pt]
\item \emph{Endpoint projection.}  Equation~\ref{eq:killed-kernel} holds with
\(\varphi_\nu(y-x)\) replaced by the density of
\(\Normal(m_1(x),\Sigma_1)\).  The survival branch has cost
\(\tfrac12(y-m_1(x))^\top\Sigma_1^{-1}(y-m_1(x))\), the cost used by
\citet{zhang2025nonequilibrium} for balanced OU bridges, and the exit branch is unchanged, so Sinkhorn again solves the projection.
\item \emph{Survivor \(x\to y\).}  The state law is Gaussian with mean
\(m_t(x)+\Sigma_t\Phi_{1-t}^\top\Sigma_1^{-1}(y-m_1(x))\) and covariance
\(\Sigma_t-\Sigma_t\Phi_{1-t}^\top\Sigma_1^{-1}\Phi_{1-t}\Sigma_t\).  The
drift is
\(b^{xy}_t(z)=Az+a+\nu\Phi_{1-t}^\top\Sigma_{1-t}^{-1}\big(y-m_{1-t}(z)\big)\),
and the exit rate is zero.
\item \emph{Exit \(x\to\coffin\).}  Before exit the particle follows the
unconditioned reference, with state law \(\Normal(m_t(x),\Sigma_t)\) and
drift \(Az+a\), and the exit rate is \(\kappa/(1-e^{-\kappa(1-t)})\) as
before.
\item \emph{Entry \(\coffin\to y\) on a growing interval.}  Reversing the
exit component of the backward problem gives, after entry, the state law
\(\Normal(m_{1-t}(y),\Sigma_{1-t})\) and the drift
\(b^{\coffin y}_t(z)=-(Az+a)-\nu\Sigma_{1-t}^{-1}\big(z-m_{1-t}(y)\big)\), and the entry clock is unchanged.  Survivors on a growing interval follow the
time reversal of the backward bridge, again a Gaussian process with an
explicit drift.
\end{itemize}

With \(A=0\) and \(a=0\), \(\Sigma_t=\nu tI\) and these expressions reduce
to Theorem~\ref{thm:reduced-lift}.  The sampling laws and regression
targets of Section~\ref{sec:population-recovery} change only through
these formulas, and the argument of
Theorem~\ref{thm:population-consistency}, which identifies the regression
optima with the Markovian projection of the bridge mixture, carries over
under the same regularity and uniqueness conditions.  For \(A\neq0\), the
reversed drifts on a growing interval differ from \(Az+a\), because the
time reversal of a linear-drift reference is a different reference.  We
checked the three drift formulas by simulating them with a fine Euler
scheme (\(\Delta t=2\times10^{-4}\), \(2\times10^4\) paths) for a
non-symmetric \(2\times2\) matrix \(A\).  The simulated means and
covariances at \(t\in\{0.3,0.7\}\) agree with the stated Gaussian laws to
within \(0.0042\), about one Monte Carlo standard error of a mean, and the
paths reach their endpoints.  The extension costs a few \(d\times d\)
matrix exponentials and inverses per time point.

\subsection{Which exit clocks keep the construction in closed form}
\label{app:clock-requirements}

The construction uses three properties of the exit clock, namely independence
of the motion, the Markov property, and closed-form conditioning on
\(\{\tau\le1\}\).  It does not require a constant rate.

\paragraph{Waiting times measured from the interval start.}
Let \(\tau\) be independent of the motion, measured from the start of the
interval as in \eqref{eq:reference-normalized}, with a continuous positive
density \(f\) on \([0,1]\) and survival function \(S\), with \(S(1)>0\).  Then \(\tau\) is
the first event of a Poisson clock with rate \(\lambda(t)=f(t)/S(t)\),
because \(S(t)=\exp(-\int_0^t\lambda)\).  Writing
\(\kappa=\int_0^1\lambda\) and \(c=\lambda/\kappa\) reduces this case to
Proposition~\ref{prop:time-only-event-clock}, where the endpoint projection
uses the survival probability \(S(1)=e^{-\kappa}\) and the conditional
exit density and active fraction follow
\eqref{eq:general-event-clock}.  Any such waiting-time law is thus an
exponential clock in a deterministically changed time, and the Mouse experiment uses one.

\paragraph{What is excluded.}
A hazard that depends on the state, \(\lambda(t,x)\), couples the clock to
the motion.  The survival branch of the endpoint kernel then carries the
Feynman--Kac factor
\(\E[\exp(-\int_0^1\lambda(t,X_t)\,\dd t)\mid X_0=x,X_1=y]\), and the
conditioned survivor paths are no longer reference bridges, so the
closed forms used here generally do not apply, and computing the projection
and targets typically requires simulation or a PDE solve.  A hazard that
depends on a hidden per-particle age, such as cell-cycle phase, makes the
reference non-Markov in the observed state.  The Markovian projection of
Theorem~\ref{thm:population-consistency} then matches the active
marginals but not the path law, and the age would have to be added to the
state.

\paragraph{Time reversal.}
The time reversal of a Markov process is Markov.  With a time-only clock,
the reversed exit component enters at the time-only rate
\(g\,\frac{\dd}{\dd t}s_\kappa(1-t)\) of Section~\ref{sec:entrance}, and an
entering particle needs no information about its future.  Memory appears
only through the age dependence excluded above, in which case neither the
forward nor the reversed process is Markov in the observed state.  The
requirement is therefore that the clock be Markov and independent of the motion.  Given these, its rate may be any positive function of time.
\end{revised}

\FloatBarrier
\section{Numerical implementation}
\label{app:implementation}

Algorithm~\ref{alg:elysium-detailed} expands the training procedure of
Section~\ref{sec:method}, and settings for each dataset are in
Appendix~\ref{app:population-training-settings}.

\begin{algorithm}[htbp]
  \caption{Expanded simulation-free training procedure}
  \label{alg:elysium-detailed}
  \small
  \begin{algorithmic}[1]
    \Require finite snapshots \(\{(t_i,\mu_i)\}_{i=0}^K\) and interval
      parameters \(\{\nu_i,\kappa_i\}\)
    \Statex \textbf{Construct the target bridge}
    \For{each adjacent interval \(i\)}
      \State orient the two snapshots from larger mass to smaller mass
      \State build the endpoint reference for killed Brownian motion using
        Equation~\ref{eq:killed-kernel}
      \State solve the endpoint coupling in
        Equation~\ref{eq:abstract-endpoint-projection}
      \State transpose the coupling if the interval grows chronologically
    \EndFor
    \Statex \textbf{Learn motion and exit}
    \Repeat
      \State sample an interval, normalized time, and live bridge component
      \State draw its state, target drift, and target exit rate from
        the conditional laws in Section~\ref{sec:population-recovery}
      \State update the drift and exit-rate networks with
        Equation~\ref{eq:training-loss}
    \Until{the learning budget is exhausted}
    \Statex \textbf{Learn entry locations}
    \For{each growing interval}
      \State fit the entry flow to live endpoints paired with the cemetery using
        Equation~\ref{eq:entrance-flow-matching}
    \EndFor
    \Ensure drift network, exit-rate network, entry flows, and interval parameters
  \end{algorithmic}
\end{algorithm}

\FloatBarrier
\subsection{Endpoint measures and coupling construction}
\label{app:endpoint-construction}
Each sampled cell carries the same weight, and snapshot mass is its cell count
relative to the initial count.  Snapshots larger than a fixed cap (2,048 to 4,096 cells, depending on the
dataset) are subsampled uniformly with a fixed seed, only to construct the
endpoint plan.  Each snapshot is subsampled once and reused by both adjacent
intervals, and every reported metric uses the full
prediction and truth clouds.  Cemetery slack lifts all endpoint measures to
the common mass \(\max_i m_i\).  The cemetery-slack argument in
Appendix~\ref{app:proof-reduced-lift} shows that this leaves the active plan
unchanged.

\FloatBarrier
\subsection{Neural fields and simulation-free training}
All non-Gaussian models use local time within each interval, separate
networks for drift and exit rate, and joint optimization against the analytic
\((b_t,h_t)\) targets in Section~\ref{sec:population-recovery}.  The learned
\(b_\theta\) and \(h_\theta\) define the SDE with the reference diffusion coefficient.
We sample times away from interval boundaries to keep the analytic targets
numerically stable.

\FloatBarrier
\subsection{Numerical generation and population events}
\label{app:numerical-generation}
Generation divides interval rates by \(\Delta_i\) and integrates
\eqref{eq:implemented-sde} with explicit stochastic midpoint.  For
evaluation with lower variance, particle weights follow the
exponential survival probabilities implied by the learned exit rate.  Exact
discrete exit times can also be sampled from that rate and define the same
law of killed paths and distribution of exit times.  On growing intervals,
midpoint quadrature integrates the entry source, with mixture centers from the auxiliary flow.
The reference fixes the source variance, while the difference in endpoint masses
fixes the total event mass.

\FloatBarrier
\section{Experimental supplement}
\label{app:experiments}

\FloatBarrier
\subsection{Evaluation metrics}
\label{app:evaluation-metrics}

\paragraph{State geometry and population mass.}
At time \(t\), write the predicted and observed finite measures as
\[
  \widehat\mu_t=\sum_{i=1}^{m_t}w_i(t)\delta_{x_i(t)},\qquad
  \mu_t=\sum_{j=1}^{n_t}\delta_{y_j}.
\]
After separately normalizing their masses, we report exact empirical
\begin{equation}
  W_1(t)=\min_{\pi\in\Pi(a,b)}
  \sum_{ij}\pi_{ij}\lVert x_i(t)-y_j\rVert_2,\quad
  a_i=\frac{w_i(t)}{\sum_kw_k(t)},\quad b_j=\frac1{n_t}.
  \label{eq:eval-w1}
\end{equation}
POT's network simplex solves this finite problem.  Every reported \(W_1\)
uses all predicted and observed points, without resampling.

Mass is evaluated directly from the full finite measures:
\begin{equation}
  \widehat r_t=\widehat m_t/\widehat m_0,\qquad
  r_t=m_t/m_0,\qquad
  \operatorname{RME}(t)=|\widehat r_t-r_t|/r_t .
  \label{eq:eval-rme}
\end{equation}
Thus \(W_1\) measures normalized state geometry and RME measures total mass.

\paragraph{Joint evaluation of state and mass on the ground-truth benchmark.}
For the ground-truth benchmark we also report two measures of error
over time.  The integrated mass error is
\(\int_0^1|\widehat m_t-m_t|\,\dd t\).  To score mass and geometry together,
let \(D\) be the diameter of the active grid and equip the grid, extended with a cemetery state,
with
\[
 d_\coffin(x,y)=\lVert x-y\rVert_2/D,\qquad
 d_\coffin(x,\coffin)=1 .
\]
For any \(L\geq\max\{\widehat m_t,m_t\}\), define
\(\widehat\mu_t^L=\widehat\mu_t+(L-\widehat m_t)\delta_\coffin\) and
\(\mu_t^L=\mu_t+(L-m_t)\delta_\coffin\).  The joint score is
\begin{equation}
  \operatorname{FM}W_1(t)=
  \frac{W_{1,d_\coffin}(\widehat\mu_t^L,\mu_t^L)}{\overline M},
  \qquad \overline M=\max\{m_0,m_1\},
  \label{eq:eval-finite-measure-w1}
\end{equation}
where \(\overline M\) is fixed for each ground-truth case and is independent
of time and the evaluated method.  The value is
independent of \(L\), since additional common cemetery slack couples to itself at
zero cost.  Hence \(\operatorname{FM}W_1\) penalizes misplaced active mass
and incorrect active mass in one exact transport problem.

\FloatBarrier
\subsection{Recovery against independent ground truth}
\label{app:formal-recovery}

\FloatBarrier
\subsubsection{Ground-truth construction and evaluation}
\paragraph{Independent reference solver.}
We construct a killed nearest-neighbor diffusion on 81-point 1D grids and
\(17\times17\) 2D grids over \([-4,4]^d\), append an absorbing cemetery
state, and evaluate its transition kernels by dense float64
scaling-and-squaring matrix exponentials.  Log-domain IPF then solves the
augmented endpoint Schr\"odinger system.  This solver
shares no implementation with \method{}'s sampler of conditional laws.  A
second uniformization calculation checks the transition kernel, and every
generated case passes the stochasticity, semigroup, IPF, and endpoint
checks.

\paragraph{Task coverage and model selection.}
Each method receives the same two weighted endpoint clouds, their exact
grid histograms, duration, \(\nu\), and \(\kappa\).  The
remaining 39 of 41 ground-truth times are used only for evaluation.  The
benchmark contains 70 cases spanning
dimension, unimodal/multimodal/crossing geometry, mass ratio, diffusion, and
killing rate.  \method{}, USB, and WFR-FM cover all 70 with five independent
training seeds \(101,\ldots,105\).  Because UDSB-F is simulation-based and
far more expensive to train, it uses five seeds on an 18-case subset
containing all 14 primary mass-ratio cases and four crossing tails, and one
seed (101) on the remaining 52 cases.
Each method uses one configuration selected on a disjoint eight-case
development set with mass ratios \(1/3\) and \(3\), and no benchmark interior
measure participates in training or selection.

\paragraph{Evaluation.}
The primary metric is conditional \(W_1\), where at each hidden time predicted and
ground-truth active measures are separately normalized and compared by the exact
1D CDF formula or POT network simplex in 2D, then averaged across the 39
times.  RME of the active mass, integrated mass error, and
\(\operatorname{FM}W_1\) provide complementary finite-measure checks.  For
methods that generate status transitions, event-time error is the \(W_1\)
distance between exit or entry CDFs, and event-location error is grid
\(W_1\) between normalized time-integrated exit or entry measures.

\FloatBarrier
\subsubsection{Recovery of state distributions and population mass}
\begin{figure}[htbp]
  \centering
  \includegraphics[width=0.94\textwidth]{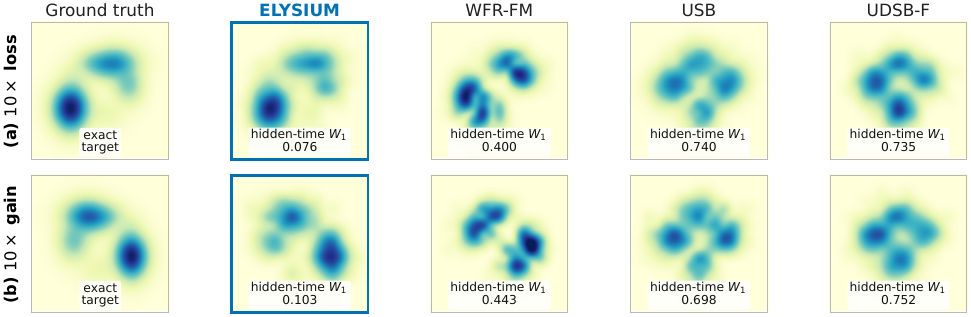}
  \caption{Hidden-distribution comparison under ten-fold population
  loss (top) and gain (bottom).  Panels show normalized distributions of the
  living population at the midpoint, and labels give state \(W_1\) averaged over
  39 hidden times.}
  \label{fig:formal-main-comparison-full}
\end{figure}

\begin{table}[htbp]
  \caption{All 70 grid cases with killing, summarized by factor.
  \method{} achieves lower conditional \(W_1\) than both USB and WFR-FM in
  every case, and the last column reports the paired mean relative reduction
  against the lower baseline in each case.  Factor groups overlap by construction.}
  \label{tab:formal-factor}
  \centering
  \small
  \setlength{\tabcolsep}{6pt}
  \begin{tabular}{llrr}
    \toprule
    Factor & Value & Cases & Mean reduction \\
    \midrule
    \rawtableinput{tables/formal_factor_rows.tex}
    \bottomrule
  \end{tabular}
\end{table}

Table~\ref{tab:formal-factor} shows that this improvement holds uniformly across
geometry, dimension, diffusivity, and event rate.  The overall paired
mean and median reductions are \(75.0\%\) and \(77.4\%\), and the smallest
casewise reduction is \(28.3\%\).  Table~\ref{tab:formal-unbalanced-metrics}
shows lower error across all 70 cases on all four metrics, including active mass, and
Table~\ref{tab:formal-primary} gives every value in the primary
mass-imbalance scan.

\begin{figure}[htbp]
  \centering
  \includegraphics[width=0.99\textwidth]{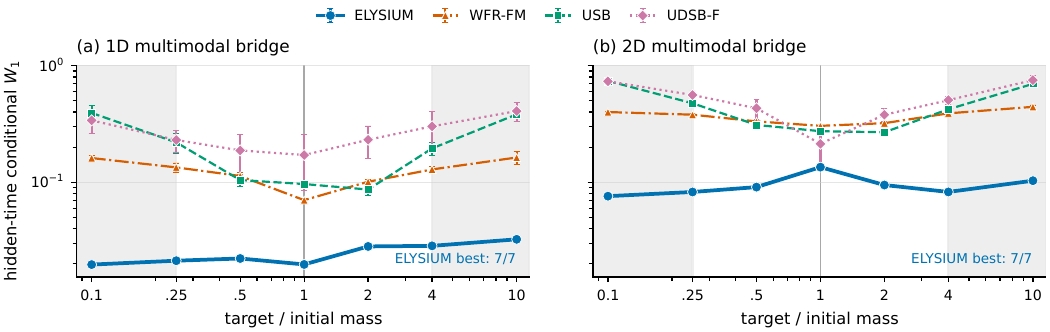}
  \caption{Primary comparison across mass ratios.  Conditional \(W_1\) is
  averaged over 39 hidden interior times, and bars show 95\% intervals across
  five fits.  Gray bands mark strong loss and gain.  UDSB-F bars use its
  18-case subset with five seeds, which contains all 14 primary cases.}
  \label{fig:formal-mass-imbalance}
\end{figure}

\begin{table}[htbp]
  \caption{Complementary scores over all 70 cases.  Each case
  first averages five fits, and ``better baseline'' selects the lower of USB and
  WFR-FM independently for each case and metric (\method{} is lower on every case).  Integrated mass error is
  \(\int_0^1|\widehat m_t-m_t|\,\dd t\).}
  \label{tab:formal-unbalanced-metrics}
  \centering
  \small
  \setlength{\tabcolsep}{5.5pt}
  \begin{tabular}{lrrr}
    \toprule
    Metric & \method{} & Better baseline & Mean reduction \\
    \midrule
    \rawtableinput{tables/formal_unbalanced_metric_rows.tex}
    \bottomrule
  \end{tabular}
\end{table}

\begin{table}[htbp]
  \caption{Conditional \(W_1\) against ground truth at hidden times
  (mean \(\pm\) sample SD over five independently trained models).  All four
  methods use the same two endpoint observations and no ground truth
  at interior times.  Lower is better.}
  \label{tab:formal-primary}
  \centering
  \scriptsize
  \setlength{\tabcolsep}{3.1pt}
  \begin{tabular}{ccrrrr}
    \toprule
    Grid & \(m_1/m_0\) & \method{} & WFR-FM & USB & UDSB-F \\
    \midrule
    \rawtableinput{tables/formal_primary_w1_rows.tex}
    \bottomrule
  \end{tabular}
\end{table}

\FloatBarrier
\subsubsection{Event recovery and simulation-based comparison}
Table~\ref{tab:formal-events} reports event recovery on the five-seed
subset.  With one paired seed on all 70 cases, conditional \(W_1\) at hidden
times is \(0.062\) for \method{} and \(0.415\) for UDSB-F, and RME of
active mass is \(0.0025\) and \(0.213\).  Across the 60 cases with nonzero
exchange, event-time \(W_1\) is \(0.0009\) versus \(0.126\), and
event-location \(W_1\) is \(0.064\) versus \(1.252\).  \rev{For
\method{}, the total mass and the aggregate event times follow from the
endpoint masses and \(\kappa\), so RME and event-time \(W_1\) check that
the generator reproduces this prior, whereas conditional and
event-location \(W_1\) measure what is learned.}

\begin{table}[!htbp]
  \caption{Event recovery on the 16 UDSB-F cases with five seeds
  and nonzero loss or gain (\method{} achieves lower error on all 16 cases).  Entries are mean (median) across paired cases, where
  each case value first averages five independent training seeds.}
  \label{tab:formal-events}
  \centering
  \small
  \setlength{\tabcolsep}{6pt}
  \begin{tabular}{lrrr}
    \toprule
    Metric & Cases & \method{} & UDSB-F \\
    \midrule
    \rawtableinput{tables/formal_event_rows.tex}
    \bottomrule
  \end{tabular}
\end{table}

\begin{revised}
\FloatBarrier
\subsubsection{Where the baselines misplace mass}
\label{app:q1-mechanism}

Two effects separate the methods in
Figure~\ref{fig:formal-main-comparison-full}.  The first is the
objective.  At \(m_1/m_0=1\) the ground truth has no net exchange, so the
differences reflect only the objective each method targets and its solver
accuracy.  WFR-FM has \(3.6\times\) and \(2.3\times\) the conditional
\(W_1\) of \method{} in 1D and 2D, and USB has \(4.9\times\) and
\(2.0\times\) (Table~\ref{tab:formal-primary}).  WFR-FM targets a
deterministic WFR geodesic and USB a branching bridge approximated by a WFR
coupling, so neither reproduces the entropic bridge of a Brownian
reference even without exchange.  The second effect is the treatment of
mass change.  At \(m_1/m_0=0.1\) the ratios grow to \(8.2\times\) and
\(5.3\times\) for WFR-FM and to \(20.0\times\) and \(9.7\times\) for USB,
and at \(m_1/m_0=10\) they are \(5.0\times\) and \(4.3\times\) for WFR-FM
and \(11.8\times\) and \(6.8\times\) for USB.

Table~\ref{tab:q1-mechanism} locates the second effect.  For each method,
we normalize the time-integrated net removal (loss) or creation (gain) of
mass, taken from the reaction term or event flux that the method exports,
and compare it with the ground-truth exit or entry flux by \(W_1\).  We also
record the normalized active mass at \(t=1/2\) in the grid cells that carry
the top half of the ground-truth exit or entry mass.  In the killed bridge,
mass that will exit stays near its source until it is removed, and mass
that enters appears near its destination, so at the midpoint these cells
still hold about a quarter of the normalized population.  \method{}
reproduces both quantities.  The baselines remove or create mass at a
\(W_1\) distance of \(0.66\)--\(1.67\) from the ground-truth event locations on the
\([-4,4]^2\) grid.  At the midpoint, WFR-FM retains \(49\%\) (loss) and
\(75\%\) (gain) of the ground-truth mass in these cells, USB \(19\%\) and
\(30\%\), and UDSB-F \(16\%\) and \(18\%\).  The mass that leaves these
cells early appears elsewhere, with USB and UDSB-F placing \(8.2\%\) and
\(6.8\%\) of their normalized mass outside the smallest set holding
\(95\%\) of the ground truth, compared with \(5\%\) for the truth by
construction and \(4.8\%\) for \method{}.  These are the additional modes
in Figure~\ref{fig:formal-main-comparison-full}.  The 1D cases give the
same ordering of event locations, \(0.023\)--\(0.039\) for \method{} and
\(0.34\)--\(1.19\) for the baselines.  The total-mass curve of \method{} is fixed by the reference rate
(Equation~\ref{eq:mass-curves}), so the diagnostics above concern where
mass changes, which the rate does not fix.

\begin{table}[!htbp]
  \caption{Mass-change diagnostics on the six 2D primary cases with net
  exchange (case means of five-seed averages).  ``Location'' is the \(W_1\)
  distance between normalized time-integrated net removal (loss) or creation
  (gain) and the ground-truth exit or entry flux.  ``Midpoint'' is the
  normalized active mass at \(t=1/2\) in the cells carrying the top half of
  the ground-truth exit or entry mass.  ``Off-support'' is the normalized active
  mass outside the smallest set holding \(95\%\) of the ground truth,
  averaged over hidden times.}
  \label{tab:q1-mechanism}
  \centering
  \small
  \setlength{\tabcolsep}{5pt}
  \begin{tabular}{lrrrrr}
    \toprule
    & \multicolumn{2}{c}{Loss (\(m_1/m_0<1\))} &
      \multicolumn{2}{c}{Gain (\(m_1/m_0>1\))} & \\
    \cmidrule(lr){2-3}\cmidrule(lr){4-5}
    Method & Location & Midpoint & Location & Midpoint & Off-support \\
    \midrule
    \rawtableinput{tables/rebuttal_q1_mechanism_rows.tex}
    \bottomrule
  \end{tabular}
\end{table}

\FloatBarrier
\subsubsection{UDSB-F with a larger training budget and finer time steps}
\label{app:udsb-budget}

UDSB-F receives the same reference and endpoint constraints as \method{}.
The Brownian parameter is mapped to its diffusivity as
\(g^2=\nu\,\Delta T\), and the killing rate to the exact per-step event
probability \(1-\exp(-\kappa\Delta T/N)\) for \(N\) solver steps.  The
endpoints are the same weighted clouds.  Because the released sampler
accepts only unweighted atoms, they are resampled once with a fixed seed,
with relative mass-ratio error below \(0.5\%\).  The differences from
\method{} therefore lie in how the bridge is learned, as UDSB-F alternates
simulated forward and backward half-bridges and learns a state-dependent
reweighting of its event prior.  To test whether the gap is a training or
discretization artifact, we trained UDSB-F on all 14 primary cases with
seeds \(101\)--\(103\) under three variations of the default
configuration, using two and four times as many training epochs per IPF phase or
\(240\) instead of \(120\) Euler steps.  Table~\ref{tab:udsb-budget}
compares each extended-budget run against the default run on the same case and seed.
None of these variations narrows the gap, as hidden-time \(W_1\) moves from
\(0.370\) to between \(0.377\) and \(0.401\), against \(0.060\) for
\method{}, fewer than half of the matched runs improve, and the mass and
event errors stay within \(7\%\) of the reported values.  Within the
tested budgets, more training or finer steps therefore do not close the
gap, which is consistent with errors from the simulated half-bridges and
the learned reweighting that \method{} replaces with closed forms.

\begin{table}[!htbp]
  \caption{UDSB-F on the 14 primary cases with more training or finer time
  steps (means over cases and seeds \(101\)--\(103\), with event metrics over the 12 cases with net exchange).  The first and last rows are the
  reported UDSB-F and \method{} runs with the same cases and seeds.
  ``Improved'' is the share of the 42 matched runs whose hidden-time \(W_1\)
  is lower than in the default run.}
  \label{tab:udsb-budget}
  \centering
  \small
  \setlength{\tabcolsep}{4pt}
  \begin{tabular}{lrrrrr}
    \toprule
    UDSB-F variant & \(W_1\) & Mass RME & Event time & Event location & Improved \\
    \midrule
    \rawtableinput{tables/rebuttal_udsb_budget_rows.tex}
    \bottomrule
  \end{tabular}
\end{table}

\FloatBarrier
\subsubsection{Simultaneous birth and death}
\label{app:bbm}

Snapshots identify only the net exchange (Appendix~\ref{app:scope}), so we
also test reconstruction when the ground truth has simultaneous birth and
death.  We use binary branching Brownian motion on the 81-point 1D grid
with \(\nu=0.25\) and the multimodal initial measure.  Each particle splits
at rate \(\lambda_s(x)\) and dies at rate \(\lambda_d(x)\).  Its expected
density solves
\(\partial_t\rho=\frac{\nu}{2}\Delta\rho+(\lambda_s-\lambda_d)\rho\), which
we evaluate exactly by a matrix exponential.  In \emph{neutral turnover},
\(\lambda_s=\lambda_d=1.5\).  In \emph{net growth}, \(\lambda_s=1\) and
\(\lambda_d=0.25\).  In \emph{opposite local change},
\(\lambda_s=0.2+1.2\,\sigma(2x)\) and
\(\lambda_d=0.2+1.2\,(1-\sigma(2x))\) with the logistic function
\(\sigma\), so the population grows on the right and shrinks on the left.
Snapshots at \(t\in\{0,\tfrac12,1\}\) are observed, with masses
\((1,1,1)\), \((1,1.455,2.117)\), and \((1,0.959,1.127)\), and the other 38
grid times are hidden.  Each method keeps its ground-truth benchmark configuration, is fitted
separately on the two observed intervals, and uses the reference rate
\(\kappa=1\) where one is required.  Only population marginals and mass are
scored, since individual splits and deaths are not identifiable from
snapshots.

\begin{table}[!htbp]
  \caption{Conditional \(W_1\) at the 38 hidden times (mean \(\pm\) sample SD over five training seeds, with the lowest in bold) and mean RME of active mass on branching Brownian motion with simultaneous birth and death.}
  \label{tab:bbm}
  \centering
  \small
  \setlength{\tabcolsep}{4pt}
  \begin{tabular}{lrrrr}
    \toprule
    Case & \method{} & WFR-FM & USB & UDSB-F \\
    \midrule
    \rawtableinput{tables/rebuttal_bbm_rows.tex}
    \bottomrule
  \end{tabular}
\end{table}

Under neutral turnover the expected density evolves as a pure diffusion,
so the snapshots are those of a process without events.  \method{} then has
no exits or entries and recovers the hidden marginals with the lowest
error.  Such turnover leaves the expected-density snapshots unchanged, so these observations cannot identify it.  Under net growth, \method{} also has
the lowest \(W_1\), while USB has the lowest mass error.  Under opposite
local change, one region gains mass while another loses it within the same
interval.  \method{} models each interval as a net loss or a net gain, so
it must transport mass between the two regions, and both models with a
signed local growth rate do better (\(W_1\) is \(0.0757\) for \method{},
\(0.0578\) for WFR-FM, and \(0.0452\) for USB).  A reference that allows
both exit and entry within one interval would remove this restriction, as discussed in Section~\ref{sec:discussion}.
\end{revised}

\FloatBarrier
\subsection{Population matching and held-out time points}
\label{app:population-time-courses}

\FloatBarrier
\subsubsection{Datasets and shared training settings}
\label{app:dataset-coverage}
\label{app:population-training-settings}
\label{app:population-matching-protocol}
\label{app:fitted-marginal-protocol}

Table~\ref{tab:dataset-coverage} lists every snapshot used in the WFR-FM
comparison.  Each setting uses five independently trained models.  When a fit
has multiple inference draws, they are averaged within that fit, so the
independently trained model remains the statistical unit.  Gaussian-1000D has
only two endpoints and therefore no interpolation holdout.

\begin{table}[!htbp]
  \caption{Dataset coverage.  Counts follow observation time.
  ``Holdouts'' is the number of distinct intermediate snapshots omitted in
  turn, with every applicable split repeated across five neural training seeds.}
  \label{tab:dataset-coverage}
  \centering
  \small
  \setlength{\tabcolsep}{4.2pt}
  \begin{tabular}{lrrll}
    \toprule
    Dataset & \(d\) & Snapshots & Cells per snapshot & Holdouts \\
    \midrule
    Simulation Gene & 2 & 5 & 400/442/530/690/969 & 3 \\
    Dyngen & 5 & 5 & 156/112/63/96/301 & 3 \\
    Gaussian-1000D & 1,000 & 2 & 500/1,400 & 0 \\
    EMT & 10 & 4 & 577/885/788/883 & 2 \\
    CITE-seq & 50 & 4 & 7,476/6,999/9,511/7,254 & 2 \\
    EB & 50 & 5 & 2,381/4,163/3,278/3,665/3,332 & 3 \\
    Mouse & 50 & 3 & 4,638/14,985/29,679 & 1 \\
    \bottomrule
  \end{tabular}
\end{table}

\paragraph{Shared neural training.}
Table~\ref{tab:fitted-marginal-training} gives the settings for drift and exit rate.
The neural fields take interval identity and local time as inputs and
regress the forward drift \(b^+\) and exit rate in a single training stage.
The non-Gaussian models use separate networks for the drift and the exit
rate, whereas Gaussian-1000D uses one network with two output heads.  All networks
receive a 16-dimensional interval embedding, and the exit-rate term in
\eqref{eq:training-loss} has weight \(0.1\).  A gaining interval also uses the entry
flow in Section~\ref{sec:entrance}.

\begin{table}[!htbp]
  \caption{Field settings for each dataset.  ``Snaps'' includes the initial snapshot.
  Every update draws the stated batch from each retained adjacent interval.
  The same row is used for population matching and every holdout experiment of
  that dataset.}
  \label{tab:fitted-marginal-training}
  \centering
  \small
  \setlength{\tabcolsep}{4.5pt}
  \begin{tabular}{lrrrrrr}
    \toprule
    Dataset & \(d\) & Snaps & Width & Blocks & Updates & Batch \\
    \midrule
    Simulation Gene & 2 & 5 & 512 & 6 & 6,000 & 400 \\
    Dyngen & 5 & 5 & 512 & 6 & 6,000 & 400 \\
    Gaussian-1000D & 1,000 & 2 & 512 & 5 & 6,000 & 400 \\
    EMT & 10 & 4 & 512 & 6 & 6,000 & 400 \\
    CITE-seq & 50 & 4 & 512 & 6 & 6,000 & 1,024 \\
    EB & 50 & 5 & 512 & 6 & 6,000 & 400 \\
    Mouse & 50 & 3 & 512 & 6 & 6,000 & 1,024 \\
    \bottomrule
  \end{tabular}
\end{table}

The entry flow uses five blocks with default width 256 (512 on Gaussian-1000D).
The default fit uses 3,000 updates, batch size 128
per gaining interval, and learning rate \(10^{-3}\), while Mouse uses 6,000 updates
and batch size 512.  The drift and exit rate stay fixed while the entry flow
is trained.

\paragraph{Reference parameters and clocks.}
The two clock families follow Appendix~\ref{app:event-clock-proof}.
\rev{Appendix~\ref{app:real-sensitivity} explains the role of these
parameters, gives a data-based rule for new datasets, and tests sensitivity
across the real time courses.}

\begin{table}[!htbp]
  \caption{Fixed reference and event settings for each dataset.  ``Sine
  \((h,a)\)'' denotes the positive global reference clock with harmonic
  \(h\) and amplitude \(a\), where identity has amplitude zero.}
  \label{tab:holdout-reference-settings}
  \centering
  \small
  \setlength{\tabcolsep}{4.5pt}
  \begin{tabular}{lrrll}
    \toprule
    Dataset & \(\nu\) & \(\kappa\) & Reference clock & Event clock \\
    \midrule
    Simulation Gene & .005 & 2 & identity & constant \\
    Dyngen & .001 & 2 & identity & constant \\
    Gaussian-1000D & .001 & 2 & identity & constant \\
    EMT & .005 & 9 & sine \((2,.13)\) & constant \\
    EB & .1 & 8 & sine \((2,.5)\) & constant \\
    CITE-seq & .01 & 2 & sine \((2,.5)\) & constant \\
    Mouse & .002 & 2 & identity & exponential tilt \(\beta=-2\) \\
    \bottomrule
  \end{tabular}
\end{table}

The sine clock is a deterministic positive time change of the killed
Brownian reference, and the Mouse event clock is the \(c_\beta(r)\) family in
\eqref{eq:exponential-event-clock}.  On the full observed range the sine
clock is
\[
 \psi_{a,h}(u)=u+\frac{a}{h\pi}\sin(h\pi u),\qquad
 \psi_{a,h}'(u)=1+a\cos(h\pi u)>0,
\]
and each retained interval uses the normalized restriction of this single
global clock, so the same \((h,a)\) applies when an interior snapshot is
omitted.  Training seeds are \(101,\ldots,105\) for every dataset, and
evaluation uses a fixed inference seed.

\paragraph{Baseline implementations and uncertainty.}
Population-matching baselines are the published WFR-FM and USB results in raw
benchmark coordinates with exact full-support \(W_1\).  SF-EUOT (EOT) and
UDSB-F are evaluated with five independently trained models each.  Because
the SF-EUOT release has no implementation for vector data on this benchmark,
we use an independent implementation of its EOT specialization with
the original paper's fixed settings for vector data, which preserves
probability mass and is evaluated only on state distributions.
UDSB-F uses the released solver and its fixed training schedule.  Holdout
WFR-FM and USB results come from their official implementations under the
same evaluator.  Every method generates from the initial population only and
passes its generated population to the next interval, with no reset to an
intermediate observation.

\FloatBarrier
\subsubsection{Population-matching protocol and results}
\label{app:fitted-marginal-results}

The model is fitted to every observed population.
Table~\ref{tab:all-baselines-fitted} summarizes the dataset means, while
Tables~\ref{tab:fitted-marginal-per-time} and~\ref{tab:evaluated-baselines-per-time}
compare the methods at every observed time after the initial snapshot.

\method{} achieves the lowest dataset-mean \(W_1\) across observed times among all compared methods on every dataset except Dyngen, where WFR-FM is slightly lower (Table~\ref{tab:all-baselines-fitted}), and improves on both WFR-FM and USB across most individual snapshots (Table~\ref{tab:fitted-marginal-per-time}).

\begin{table}[htbp]
  \caption{Mean population-matching \(W_1\) for all methods.
  \method{}, SF-EUOT (EOT), and UDSB-F are evaluated over five seeds in
  this work, whereas WFR-FM and USB values are taken from their papers.  The middle
  block is reported by WFR-FM
  \citep{peng2026wfrfm}, WFR-MFM is reported in its paper
  \citep{wang2026wfrmfm}, and BranchSBM in USB \citep{ying2026beyond}.
  A dash means that the source paper did not report the dataset.}
  \label{tab:all-baselines-fitted}
  \centering
  \scriptsize
  \resizebox{\textwidth}{!}{%
  \begin{tabular}{lrrrrrrr}
    \toprule
    Method & Sim.\ Gene & Dyngen & Gaussian & EMT & EB & CITE-seq & Mouse \\
    \midrule
    \rawtableinput{tables/all_baseline_nonholdout_rows.tex}
    \bottomrule
  \end{tabular}}
\end{table}

\begin{table}[htbp]
  \caption{All constrained population marginals.  \method{} entries are mean
  \(\pm\) sample SD over five independent neural fits.  RME is the
  corresponding mean relative error in total mass.  Bold marks the lower per-time
  \(W_1\).}
  \label{tab:fitted-marginal-per-time}
  \centering
  \scriptsize
  \setlength{\tabcolsep}{4.2pt}
  \begin{tabular}{llrrrr}
    \toprule
    Dataset & Time & WFR-FM \(W_1\) & USB \(W_1\) &
      \method{} \(W_1\) & \method{} RME \\
    \midrule
    \rawtableinput{tables/fitted_marginal_per_time_rows.tex}
    \bottomrule
  \end{tabular}
\end{table}

\begin{table}[htbp]
  \caption{Per-time results for SF-EUOT (EOT) and UDSB-F.  Entries are
  mean \(\pm\) sample SD over five independently
  trained models.}
  \label{tab:evaluated-baselines-per-time}
  \centering
  \scriptsize
  \setlength{\tabcolsep}{5pt}
  \begin{tabular}{llrr}
    \toprule
    Dataset & Time & SF-EUOT (EOT) & UDSB-F \\
    \midrule
    \rawtableinput{tables/evaluated_baselines_per_time_rows.tex}
    \bottomrule
  \end{tabular}
\end{table}

\FloatBarrier
\subsubsection{Protocol and results for omitted time points}
\label{app:real-holdout-protocol}

Every interior time listed in Table~\ref{tab:dataset-coverage} is omitted in
turn.  The endpoint plan, drift, exit rate, and entry flow
are learned using only the retained snapshots, and the generator propagates
the initial population continuously to the omitted time.  Standardization statistics are fitted only to retained
snapshots and inverted before \eqref{eq:eval-w1}.  The omitted population's
states and count are used only for evaluation.

Holdout refits use the configuration in
Appendix~\ref{app:population-training-settings}, and an entry flow is fitted
only when a retained interval gains mass.  Inference uses 100 stochastic midpoint steps per interval and a
fixed inference seed.

\begin{table}[!htbp]
  \caption{The eight individual real-data interpolation results, with dataset
  averages in Table~\ref{tab:all-baselines-holdout}.  Each omitted
  snapshot is predicted by a model
  trained on the remaining times.  Entries are mean \(\pm\) sample SD over
  five fits.}
  \label{tab:real-holdout-per-time}
  \centering
  \scriptsize
  \setlength{\tabcolsep}{2.6pt}
  \begin{tabular}{llrrrr}
    \toprule
    Dataset & Omit & WFR-FM \(W_1\) & USB \(W_1\) & UDSB-F \(W_1\) &
      \method{} \(W_1\) \\
    \midrule
    \rawtableinput{tables/holdout_per_split_rows.tex}
    \bottomrule
  \end{tabular}
\end{table}

\begin{table}[!htbp]
  \caption{Real-data holdout \(W_1\), averaged over omitted snapshots.
  \method{}, WFR-FM, USB, and UDSB-F were run for five seeds.  The middle block is reported in WFR-FM
  \citep{peng2026wfrfm}, WFR-MFM is reported in its Appendix Table~3
  \citep{wang2026wfrmfm}, and BranchSBM in USB's Main Table~4
  \citep{ying2026beyond}.  Lower is better.}
  \label{tab:all-baselines-holdout}
  \centering
  \small
  \setlength{\tabcolsep}{6pt}
  \begin{tabular}{lrrrr}
    \toprule
    Method & EMT & EB & CITE-seq & Mouse \\
    \midrule
    \rawtableinput{tables/all_baseline_holdout_rows.tex}
    \bottomrule
  \end{tabular}
\end{table}

\method{} achieves lower \(W_1\) than UDSB-F at every omitted snapshot and improves on WFR-FM and USB across nearly all individual holdouts (Table~\ref{tab:real-holdout-per-time}).  Averaged within each dataset,
it outperforms both principal baselines on EMT, EB, and Mouse, while WFR-FM achieves lower error on
CITE-seq, and Table~\ref{tab:all-baselines-holdout} extends this comparison
to the other reported methods.

\FloatBarrier
\subsection{4i drug-response dynamics}
\label{app:drug-protocol}

\paragraph{Data and evaluation split.}
The drugs are administered at time zero.  The 4i Trametinib+Erlotinib condition
then has \(2452/3322/1841\) cells at 8/24/48h and 50 morphology and
protein intensity features
\citep{bunne2021learning,pariset2023unbalanced}.  The public 80/20 split
contains \(1962/2684/1479\) training cells and \(490/638/362\) evaluation
cells.  We retain this split and the original 50-dimensional coordinates for
every method.  The training portion of all three observations and the
corresponding cell counts are available during model construction, whereas the
evaluation cells are not used for fitting.  All generated trajectories start from the 8h
population and continue through 24h to 48h without an observation reset.

\paragraph{Model configuration and continuous generation.}
The \method{} network has width 512, six residual blocks, eight Fourier
frequencies, and a 16-dimensional interval embedding.  It uses the same
single-stage parameterization as the other non-Gaussian experiments, with one
joint fit of the forward drift and exit rate over 6,000 updates using separate
drift and exit-rate networks.  The 8--24h growth interval also
trains the standard entry flow described in
Section~\ref{sec:entrance}, using width 256, five residual blocks, and 3,000
updates, while the 24--48h interval uses exit.  The reference uses \(\nu=.01\) and
\(\kappa=2\).  The five training seeds are \(23,31,47,59,71\).  Each fit starts
from a paired 512-cell 8h training cloud and uses one continuous
SDE trajectory with the reference diffusion coefficient and 100 stochastic midpoint steps per
interval.  The support is not capped after initialization, allowing newly
generated entries to remain in the population.

\paragraph{Paired baseline protocol.}
For WFR-FM and USB, we train their released model implementations on all three
observed snapshots using fixed 50-dimensional configurations.  Each method is
then propagated from 8h to 48h.  To give UDSB-F the
same three observed populations, the baseline fitted to all snapshots
trains the released solver separately on the adjacent 8--24h and 24--48h
intervals, retaining its architecture, training schedule, and the dependence
of its event prior on state.  During generation, the second interval starts from the
exact terminal particle positions and active/dead indicators produced by the
first.  All four methods use
the same five paired 512-cell initial clouds and sampling seeds, and are scored
with the same exact \(W_1\) evaluator in the original 50 coordinates.
\paragraph{Results.}
Table~\ref{tab:drug} gives the paired comparison at both evaluation times.
\begin{table}[!htbp]
  \caption{4i Trametinib+Erlotinib population matching in the original 50
  features on the same held-out cells.  Values are mean and sample SD over
  five independently trained models.}
  \label{tab:drug}
  \centering
  \small
  \setlength{\tabcolsep}{7pt}
  \begin{tabular}{lcc}
    \toprule
    Method & 24h \(W_1\downarrow\) & 48h \(W_1\downarrow\) \\
    \midrule
    \rawtableinput{tables/drug_w1_rows.tex}
    \bottomrule
  \end{tabular}
\end{table}
\FloatBarrier
\subsection{Sensitivity to bridge variance and killing rate}
\label{app:simulation-gene-sensitivity}

We assess sensitivity on the Simulation Gene holdouts over a finite grid of
Brownian bridge variance
\(\nu\in\{0.001,0.005,0.02\}\) and constant killing rate
\(\kappa\in\{0.5,2,8\}\).  Every pair is refitted for all three omitted
interior snapshots with training seeds \(101,\ldots,105\), while the architecture,
optimizer, observed masses, and direct parameterization of the forward drift
remain fixed.  All settings use the identity reference clock and an event rate
that is constant in time.  This auxiliary
study uses 400 stochastic midpoint steps and three inference seeds per trained model.
Inference results are first averaged for each model, after which we
report the mean and sample SD over the five independent fits.  The main
benchmark tables retain their fixed 100-step protocol.

\begin{table}[!htbp]
  \caption{Sensitivity to reference parameters on Simulation Gene at the three held-out
  interior times.  \(W_1\) entries are mean \(\pm\) sample SD over five
  independently trained models, and RME entries are the corresponding mean relative
  mass errors, whose SD across fits is zero.  Bold hyperparameters mark the
  benchmark setting.}
  \label{tab:simulation-gene-sensitivity}
  \centering
  \small
  \setlength{\tabcolsep}{2.8pt}
  \begin{tabular}{ccrrrrrr}
    \toprule
    \multicolumn{2}{c}{Hyperparameters} &
    \multicolumn{2}{c}{\(t_1\)} &
    \multicolumn{2}{c}{\(t_2\)} &
    \multicolumn{2}{c}{\(t_3\)} \\
    \cmidrule(lr){1-2}\cmidrule(lr){3-4}\cmidrule(lr){5-6}\cmidrule(lr){7-8}
    \(\nu\) & \(\kappa\) &
    \multicolumn{1}{c}{\(W_1\)} & \multicolumn{1}{c}{RME} &
    \multicolumn{1}{c}{\(W_1\)} & \multicolumn{1}{c}{RME} &
    \multicolumn{1}{c}{\(W_1\)} & \multicolumn{1}{c}{RME} \\
    \midrule
    \rawtableinput{tables/simulation_gene_sensitivity_rows.tex}
    \bottomrule
  \end{tabular}
\end{table}

Across this finite grid, changing \(\nu\) at fixed \(\kappa\) has a
comparatively modest effect on holdout \(W_1\), whereas \(\kappa=8\)
substantially degrades both state and mass interpolation.  The benchmark pair
\((\nu,\kappa)=(0.005,2)\) has the nominally lowest holdout macro \(W_1\), but
its difference from \((0.001,2)\) is smaller than the SD across training seeds.
At fixed \(\kappa\), RME is unchanged across \(\nu\), consistent with
\eqref{eq:mass-curves}, since \(\nu\) controls the spatial reference whereas
\(\kappa\) controls event timing within each interval.  No single pair is best at
every omitted time.

\begin{revised}
\FloatBarrier
\subsection{Reference settings and masses on the real time courses}
\label{app:real-sensitivity}

\paragraph{Role of the parameters.}
The coupling is solved in isotropically standardized coordinates
(Appendix~\ref{app:endpoint-construction}), so \(\nu\) is the reference
variance per coordinate over one normalized interval in units of the data's
RMS scale.  The killing rate \(\kappa\) does not enter the endpoint coupling
(constant block factors are absorbed by Sinkhorn) and affects only when
events occur within an interval.  By Equation~\ref{eq:mass-curves},
\(\kappa=|\log(m_1/m_0)|\) makes the total mass the constant-rate
exponential \(m_0(m_1/m_0)^t\).  Smaller \(\kappa\) spreads events more
evenly in time, and larger \(\kappa\) shifts exits toward the start of a
decreasing interval and entries toward the end of a growing one.  The
clocks reshape this timing.  The values in
Table~\ref{tab:holdout-reference-settings} were set per dataset, and each dataset uses one row for population matching and
all holdouts.

\paragraph{Shared default and data-based rule on a new dataset.}
For a new dataset, one can either use the shared default (\(\nu=0.005\),
\(\kappa=2\), identity clocks, as on Simulation Gene) or set
\((\nu,\kappa)\) directly from the observed training snapshots by setting
\(\kappa\) to the mean of \(|\log(m_{k+1}/m_k)|\) over adjacent
training snapshots (matching the constant-rate value in
Equation~\ref{eq:mass-curves}), \(\nu\) is \(1\%\) of the mean squared
transport displacement per coordinate between adjacent training snapshots
in standardized coordinates, and the clocks are the identity.  This rule
gives \(\kappa\) between \(0.05\) and \(1.86\) and \(\nu\) between
\(0.004\) and \(0.019\).  Table~\ref{tab:shared-default} refits every
dataset across five training seeds under both choices and
compares them against five-seed baseline fits under the per-dataset settings of Table~\ref{tab:holdout-reference-settings} (using the same
architecture, optimizer, training seeds, and 100-step inference protocol of
Appendix~\ref{app:population-training-settings}, matching the main-table \method{} means to within \(3.1\%\)).
Relative to the per-dataset setting, population-matching \(W_1\) changes by
\(-12.3\%\) to \(+3.5\%\) under the shared default and by \(-1.3\%\) to
\(+7.1\%\) under the data-based rule, and both keep the same ranking against
WFR-FM and USB on all seven datasets (leading on six of seven, with Dyngen
the exception).  Holdout \(W_1\) changes by \(-0.2\%\) to \(+5.1\%\) under
the shared default and by \(+0.7\%\) to \(+6.0\%\) under the rule, and 4i
\(W_1\) changes by at most \(1.8\%\).

\begin{table}[!htbp]
  \caption{The shared setting (\(\nu=0.005\), \(\kappa=2\), identity
  clocks) and the data-based rule, compared with the per-dataset
  settings of Table~\ref{tab:holdout-reference-settings}.  The top block gives population-matching \(W_1\), the middle block holdout \(W_1\) averaged over omitted snapshots, and the bottom block 4i \(W_1\) on the held-out cells.  Entries are mean\(\pm\)sample SD over five fits, and ``Change'' is the relative difference from the per-dataset mean.}
  \label{tab:shared-default}
  \centering
  \small
  \setlength{\tabcolsep}{3.5pt}
  \begin{tabular}{lrrrrr}
    \toprule
    & & \multicolumn{2}{c}{Shared setting} & \multicolumn{2}{c}{Data-based rule} \\
    \cmidrule(lr){3-4}\cmidrule(lr){5-6}
    Dataset & Per-dataset & \(W_1\) & Change & \(W_1\) & Change \\
    \midrule
    \rawtableinput{tables/rebuttal_shared_default_rows.tex}
    \bottomrule
  \end{tabular}
\end{table}

\paragraph{Endpoint masses.}
The masses enter only through the endpoint constraints and the mass curve
in Equation~\ref{eq:mass-curves}.  We refit every real holdout with three
alternatives to the observed relative counts \(m_k\) (equal masses, and
\(m_k^{\alpha}\) with \(\alpha=\tfrac12\) or \(2\), which halve or double
every log mass ratio) using three training seeds.  On EMT, EB, and CITE-seq
the holdout \(W_1\) changes by at most \(2.3\%\)
(Table~\ref{tab:mass-perturbation}), so the state error there depends
little on the counts.  Mouse, whose profiled population more than triples
between the first two snapshots, is the exception, as every perturbation
raises \(W_1\) by \(8\)--\(12\%\), showing that its observed counts carry information that the generator uses.

\begin{table}[!htbp]
  \caption{Holdout \(W_1\) with perturbed endpoint masses, averaged over
  omitted snapshots.  Rows other than ``Observed'' are mean\(\pm\)sample SD
  over three fits, and ``Observed'' is the five-seed baseline fit from
  Table~\ref{tab:shared-default}.}
  \label{tab:mass-perturbation}
  \centering
  \small
  \setlength{\tabcolsep}{5pt}
  \begin{tabular}{lrrrr}
    \toprule
    Masses & EMT & EB & CITE-seq & Mouse \\
    \midrule
    \rawtableinput{tables/rebuttal_mass_rows.tex}
    \bottomrule
  \end{tabular}
\end{table}

\paragraph{Grid over \(\nu\), \(\kappa\), and the clock.}
Figure~\ref{fig:real-grid} refits all eight holdouts for
\(\nu\in\{0.001,0.005,0.02,0.1\}\) and \(\kappa\in\{0.5,2,8\}\), with
identity clocks or with the dataset's clock, using three training seeds.
Across the entire 24-point grid, holdout \(W_1\) stays within \(3.2\%\) of
the reported setting on EMT (\(0.2983\)--\(0.3079\)), within \(4.2\%\) on
CITE-seq (\(37.17\)--\(38.92\)), and within \(6.9\%\) on EB
(\(9.77\)--\(10.45\), or within \(4.9\%\) for \(\kappa\in\{2,8\}\)).  At
fixed \(\kappa\) and clock, varying \(\nu\) over two orders of magnitude
changes \(W_1\) by at most \(0.9\%\) on EMT, \(2.7\%\) on CITE-seq, and
\(3.1\%\) on EB.  Mouse is sensitive mainly to \(\kappa\).  Across
\(\kappa\in\{0.5,2\}\) and \(\nu\le0.02\) around its reported setting,
\(W_1\) is \(6.29\)--\(7.12\) (\(-3.7\%\) to \(+8.9\%\) relative to the
reported \(6.54\)), whereas \(\kappa=8\) raises \(W_1\) to \(8.6\)--\(9.7\)
because the held-out interval has a \(6.4\times\) mass increase and
Equation~\ref{eq:mass-curves} with \(\kappa=8\) delays \(98\%\) of that
increase past the midpoint, whereas the rule gives \(\kappa=1.86\) for this
holdout.  The effect of \(\kappa\) therefore grows with the interval mass
ratio.  All statements in this section concern \(W_1\) of the predicted states, and the total mass at an omitted time follows
Equation~\ref{eq:mass-curves}.

\begin{figure}[!htbp]
  \centering
  \includegraphics[width=\linewidth]{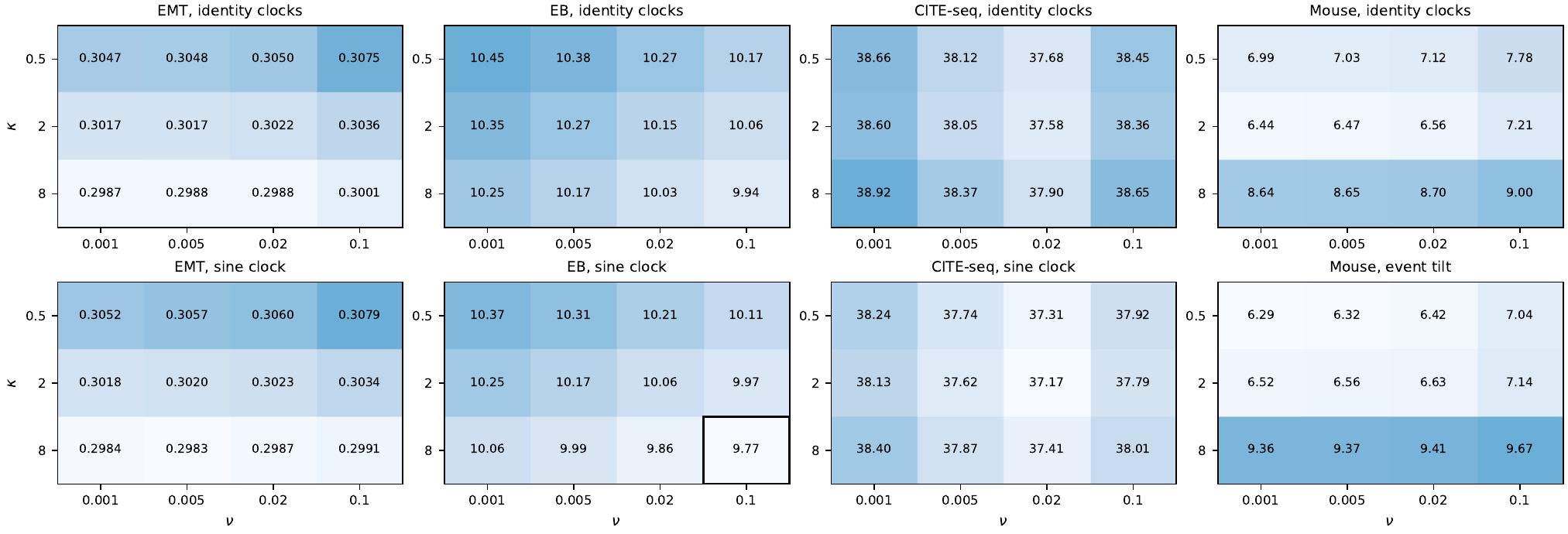}
  \caption{Holdout \(W_1\) over the reference grid, averaged over omitted
  snapshots and three training seeds.  The top row uses identity reference and event clocks, and the bottom row the clock of
  Table~\ref{tab:holdout-reference-settings}.  Shading is scaled within
  each dataset from its grid minimum (light) to maximum (dark).  The
  outlined EB cell is the reported setting, and the reported EMT, CITE-seq, and Mouse settings lie off the grid.}
  \label{fig:real-grid}
\end{figure}
\end{revised}

\FloatBarrier
\subsection{Computational cost}
\label{app:compute}

\begin{table}[!htbp]
  \caption{Training time and peak GPU memory for the
  holdout experiments.  Training times report mean\(\pm\)sample SD across fits, where
  \(n\) is the number of UDSB-F runs, with no SD for the single Mouse run.
  Peak memory for \method{} and WFR-FM is reported as mean/maximum allocated GiB.}
  \label{tab:compute-cost}
  \centering
  \footnotesize
  \setlength{\tabcolsep}{4pt}
  \begin{tabular}{lrrrrrr}
    \toprule
    & \multicolumn{2}{c}{\method{}} & \multicolumn{2}{c}{WFR-FM} &
      \multicolumn{2}{c}{UDSB-F} \\
    \cmidrule(lr){2-3}\cmidrule(lr){4-5}\cmidrule(lr){6-7}
    Dataset & Train (min) & Peak GiB & Train (min) & Peak GiB & Train (min) & \(n\) \\
    \midrule
    \rawtableinput{tables/compute_rows.tex}
    \bottomrule
  \end{tabular}
  \par\smallskip
  \begin{minipage}{\linewidth}
  \footnotesize
  \emph{Hardware and timing.} \method{} and WFR-FM use an RTX~4090, and
  their times include endpoint construction and all training stages.
  UDSB-F uses an RTX~4090~D, and its times cover training only, excluding initialization.
  \end{minipage}
\end{table}

\method{} averages
\(1.41\)--\(2.12\) minutes per fit across the six datasets, whereas
UDSB-F means for the training loop range from \(2.13\) to \(6.34\) minutes.  \method{} remains
comparable to WFR-FM in training time, with maximum allocated GPU memory
of \(0.89\)~GiB versus \(5.86\)~GiB for WFR-FM.

\begin{revised}
\subsubsection{Endpoint coupling and end-to-end cost}
\label{app:ot-cap}

The endpoint coupling is computed once per interval, before neural
training.  For snapshots of \(n_0\) and \(n_1\) cells, each Sinkhorn
iteration costs \(O(n_0n_1)\) time and memory, and the dimension enters only once, through the cost matrix.  On the three
largest datasets, the subsampling cap of
Appendix~\ref{app:endpoint-construction} bounds \(n_0\) and \(n_1\).
Table~\ref{tab:end-to-end} times every
stage of one population-matching fit per dataset, each run
on a single dedicated RTX~4090~D GPU.  The coupling takes \(2.5\)--\(14.6\)~s, at most
\(11\%\) of the total, and the whole pipeline, including 100-step SDE generation,
takes \(1.1\)--\(2.5\) minutes with at
most \(0.89\)~GiB of GPU memory.

\begin{table}[!htbp]
  \caption{Wall-clock cost of one population-matching fit per dataset on a
  dedicated RTX~4090~D GPU (seconds, except the total).  ``Drift and exit'' is the
  joint simulation-free regression, ``Entry'' the entry flow, and
  ``Generation'' SDE integration from the initial population with 100
  midpoint steps per interval.  Evaluation of the exact \(W_1\) is
  excluded.}
  \label{tab:end-to-end}
  \centering
  \small
  \setlength{\tabcolsep}{4.5pt}
  \begin{tabular}{lrrrrrr}
    \toprule
    Dataset & Coupling & Drift and exit & Entry & Generation &
      Total (min) & Peak GiB \\
    \midrule
    \rawtableinput{tables/rebuttal_end_to_end_rows.tex}
    \bottomrule
  \end{tabular}
\end{table}
\end{revised}

\FloatBarrier
\subsection{Supplementary neural SDE trajectories}
\label{app:trajectory-atlas}

\begin{figure}[!htbp]
  \centering
  \includegraphics[width=0.78\textwidth]{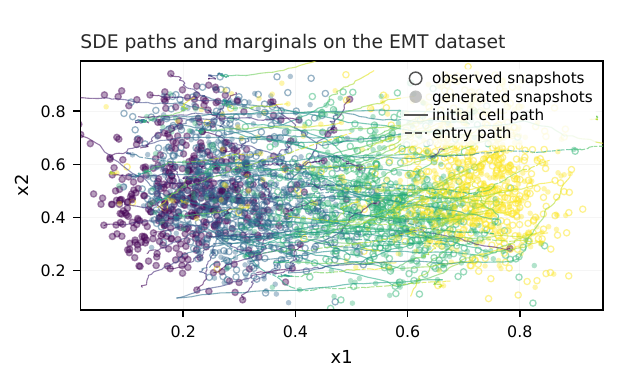}
  \caption{Individual SDE paths with observed and generated snapshot populations on EMT.}
  \label{fig:atlas-emt}
\end{figure}

\begin{figure}[!htbp]
  \centering
  \includegraphics[width=0.78\textwidth]{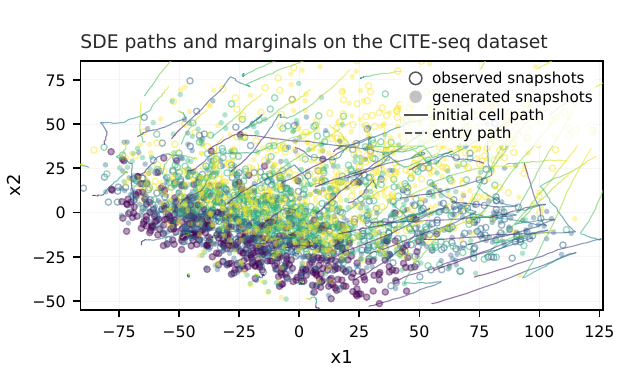}
  \caption{Individual SDE paths with observed and generated snapshot populations on CITE-seq.}
  \label{fig:atlas-cite}
\end{figure}

\begin{figure}[!htbp]
  \centering
  \includegraphics[width=0.78\textwidth]{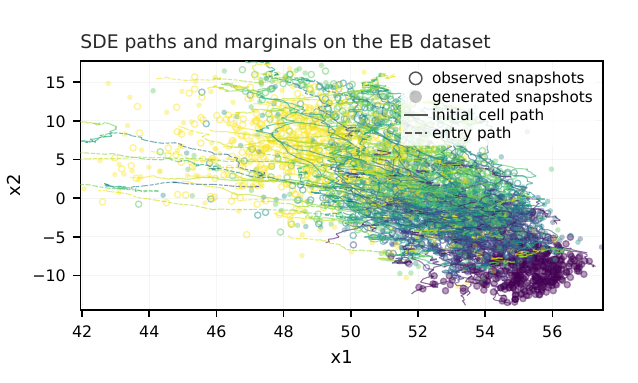}
  \caption{Individual SDE paths with observed and generated snapshot populations on EB.}
  \label{fig:atlas-eb}
\end{figure}

\begin{figure}[!htbp]
  \centering
  \includegraphics[width=0.78\textwidth]{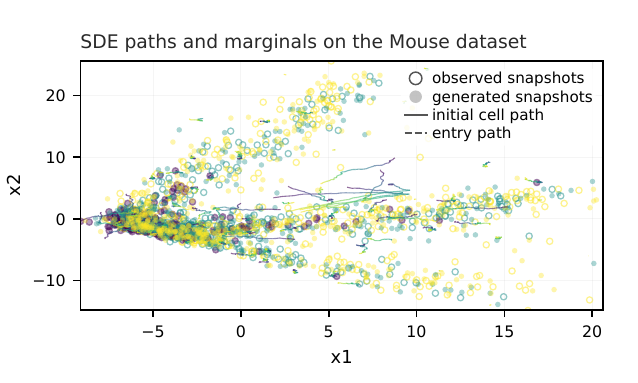}
  \caption{Individual SDE paths with observed and generated snapshot populations on Mouse.}
  \label{fig:atlas-mouse}
\end{figure}

Figures~\ref{fig:atlas-emt}--\ref{fig:atlas-mouse} complement the
population-matching results with one representative model per real dataset.
Each generated trajectory starts from the empirical initial cloud and uses the learned
drift with the reference diffusion coefficient.  State panels show the first
two benchmark coordinates without additional PCA or display transforms, while
generation and \(W_1\) evaluation use all benchmark coordinates.

Hollow and filled circles denote observed and generated populations.
Solid paths start from the initial cloud, and dashed paths are independently
generated entries.  Colors indicate time.  Each panel shows paths
of 48 initial cells and up to 24 entering cells.
Paths are model samples, not observed lineages.

\FloatBarrier
\section{Relationship to models of growth and individual events}
\label{app:scope}

\FloatBarrier
\subsection{Objectives and generative mechanisms}

\begin{table}[!htbp]
  \caption{Comparison of closely related neural formulations.  Models with
  explicit events generate stochastic exits, entries, or branching.  The final
  column indicates whether training targets the objective
  in the second column at population optimum, whereas USB presents its solver
  as an approximation to the branching SB \citep{ying2026beyond}.}
  \label{tab:method-comparison}
  \centering
  \small
  \setlength{\tabcolsep}{2.4pt}
  \begin{tabular}{lp{0.25\textwidth}cccc}
    \toprule
    Method & Objective & \shortstack{Stochastic\\paths} &
      \shortstack{Explicit\\events} & \shortstack{Simulation-free\\training} &
      \shortstack{Targets stated\\objective} \\
    \midrule
    SF-EUOT & entropic UOT control & \(\checkmark\) & \(\times\) & \(\checkmark\) & \(\checkmark\) \\
    WFR-FM & WFR dynamic action & \(\times\) & \(\times\) & \(\checkmark\) & \(\checkmark\) \\
    UDSB & killed-diffusion path KL & \(\checkmark\) & \(\checkmark\) & \(\times\) & \(\checkmark\) \\
    USB & branching path KL & \(\checkmark\) & \(\checkmark\) & \(\checkmark\) & \(\times\) \\
    \method{} & killed-diffusion path KL & \(\checkmark\) & \(\checkmark\) & \(\checkmark\) & \(\checkmark\) \\
    \bottomrule
  \end{tabular}
\end{table}

\begin{revised}
\FloatBarrier
\subsection{Relation to prior killed Schr\"odinger bridge formulations}
\label{app:inherited}

Table~\ref{tab:inherited} separates the problem formulation, which
\method{} takes from earlier work on killed Schr\"odinger bridges, from the
parts that make simulation-free training possible.  In particular, the
objective solved by \method{} is the same killed-diffusion path KL as in
\citet{chen2022most} and UDSB, and the difference from UDSB lies in how the
bridge is learned.

\begin{table}[!htbp]
  \caption{Components of \method{} taken from earlier work and components
  introduced here.}
  \label{tab:inherited}
  \centering
  \small
  \setlength{\tabcolsep}{4pt}
  \begin{tabular}{p{0.5\textwidth}p{0.2\textwidth}p{0.2\textwidth}}
    \toprule
    Component & Status & Where \\
    \midrule
    Killed Brownian reference, path-KL objective with unbalanced marginals,
    cemetery state, reverse-time entry
      & from \citet{chen2022most,pariset2023unbalanced}
      & Sec.~\ref{sec:abstract-usb} \\
    Reduction to an endpoint problem and its Sinkhorn solution
      & from \citet{leonard2014survey,chen2022most}
      & Sec.~\ref{sec:cemetery-lift} \\
    \midrule
    Closed-form survival, exit, and entry components of the optimal bridge
    for each endpoint pair (Theorem~\ref{thm:reduced-lift})
      & new & Sec.~\ref{sec:components} \\
    Simulation-free regression targets for drift (squared loss) and exit
    rate (Poisson loss), and the entry flow with its time-only clock
      & new & Secs.~\ref{sec:population-recovery}--\ref{sec:entrance} \\
    Recovery of the bridge by the regression optima and empirical
    consistency (Theorem~\ref{thm:population-consistency})
      & new & Sec.~\ref{sec:consistency}, App.~\ref{app:proofs} \\
    Time-dependent clocks, multi-marginal separation, linear-drift
    references
      & new & App.~\ref{app:event-clock-proof}, \ref{app:static-structure}--\ref{app:clock-requirements} \\
    \bottomrule
  \end{tabular}
\end{table}
\end{revised}

\FloatBarrier
\subsection{Identifiability and the role of the reference}
\label{app:identifiability}

WFR and other models of growth describe changing populations through
a signed reaction field in the continuity equation
\citep{chizat2018scaling,liero2018optimal,peng2026wfrfm}.  This field identifies
the net local change in population density.  To see the resulting ambiguity,
write a population equation with per-capita birth and death rates \(\beta\) and
\(\delta\):
\[
  \partial_t\rho + \nabla\!\cdot(\rho v)
  = (\beta-\delta)\rho.
\]
The snapshots constrain only the difference \(g=\beta-\delta\).  For every
nonnegative field \(c(t,x)\), the pair \((\beta+c,\delta+c)\) produces the same
equation and hence the same population marginals.  A signed growth field thus
gives a concise macroscopic description, but specifying individual events
requires an additional stochastic model.

Branching SBs instead choose a branching process as the reference law
\citep{baradat2021branching,tang2025branchsbm,ying2026beyond}.  Their events
have a parent--offspring interpretation and naturally represent division.
Here too, anonymous snapshots identify less than an experiment that tracks lineage,
since if \(q_k(t,x)\) is the rate of producing \(k\) descendants, the population
equation depends only on the first moment
\(\sum_k (k-1)q_k(t,x)\), so distinct offspring laws can yield the same
population evolution.  A branching reference selects among these possibilities
through its prior, whereas branch-specific models additionally require a partition of
terminal states into branches.  USB learns a simulation-free approximation to
this branching objective and uses the inferred fields to simulate division
and death \citep{ying2026beyond}.

\method{} uses a killed diffusion with a cemetery state.  Every active
path either persists between observed states or exits at a stochastic time, and
reading the same bridge backward gives anonymous entry.  The path-space
entropy objective jointly determines the survivor coupling, state motion,
and distributions of event times and entry locations.  Consequently, the reaction
terms in Equation~\ref{eq:fixed-pool-continuity} are realized by a specific
stochastic event process, so the learned dynamics describe individual histories
without requiring lineage observations.

\begin{revised}
\FloatBarrier
\subsection{Biological reading of exits, entries, and masses}
\label{app:biology}

\paragraph{Matching.}
Cells measured at different times are different cells, so the coupling
\(\pi^*\) describes the most likely way, under the reference, for the population at \(t_i\) to
become the population at \(t_{i+1}\), where \(\pi^*(x,y)\) is the mass in state
\(x\) that the model sends to state \(y\) and \(\pi^*(x,\coffin)\) is the
mass that leaves the population.  The learned generator turns this into a
stochastic process for individual cells, that is, a model of individual
histories that is consistent with the snapshots, which themselves contain
no histories.

\paragraph{Exits.}
An exit removes a cell from the observed population.  Depending on the
system, this may be death, differentiation out of the profiled
compartment, emigration from the sampled tissue, or loss before
measurement.  The model represents all of these by one mechanism, so the
exit rate \(h(t,x)\) estimates when and in which states the population is
depleted.  In the 4i experiment, the population
contracts between 24 and 48 hours of combined inhibition, which earlier
work attributes to drug-induced cell death \citep{pariset2023unbalanced}.

\paragraph{Entries.}
An entry adds a cell.  In a proliferating population, added mass comes
from division, and in tissues also from influx or from differentiation into
the compartment.  Sampling contributes as well, because when a state is captured or
sorted at a higher rate at the later time, or the sampled tissue region
changes, the profiled population gains mass in that state.  The reference has no parent--offspring structure, so an entry at state \(y\) and time \(t\) means that the population gains mass
near \(y\) at time \(t\).  The entry source \(\eta_t\) of
Equation~\ref{eq:entry-source} depends only on time and the coupling, so
mass can enter states where the profiled population currently has little
or no mass.  A net growth rate \(\gamma\) enters the continuity equation as
\(\gamma\rho_t\) and creates mass in proportion to the mass already present, so
adding mass at a sparsely populated state requires either a large local
growth rate or first transporting cells there.  A division at
state \(z\) produces two cells near \(z\), which at the population level is
an entry near \(z\).  A branching reference, which ties offspring to parents, or
lineage data would resolve division itself
\citep{baradat2021branching,weinreb2020lineage}.

\paragraph{Masses.}
As in WFR-FM and its benchmark \citep{peng2026wfrfm}, the mass of each
snapshot is its number of profiled cells relative to the first snapshot.
For the simulated datasets, these counts are part of the benchmark design.
For the single-cell RNA-seq time courses (EMT, EB, CITE-seq, and Mouse),
the number of profiled cells also depends on how many cells were loaded,
captured, and retained after quality control at each time, so the masses
are relative abundances of the profiled population, and exits and
entries describe changes of this profiled population.  Technical gains and losses of this kind enter the model as entries and
exits in the same way as biological events.  A mass-conserving model
would explain such state-dependent changes by transport, so that a state
captured more often at the later time appears as cells migrating toward
it.  In the cemetery lift, the masses fix how much mass exits or enters in
each interval.  Within that amount, surviving mass pays a cost that grows with the
distance travelled, while exiting or entering mass pays a cost independent
of position (Section~\ref{sec:cemetery-lift}), so mass with no nearby
counterpart in the other snapshot is preferentially assigned to exit or
entry, and motion is inferred only for the matched mass.  The method takes
the masses as input, so external cell counts or capture-efficiency estimates
can replace relative counts directly, and Appendix~\ref{app:real-sensitivity}
evaluates sensitivity to perturbing these endpoint masses across the real holdouts.
\end{revised}

\end{document}